\documentclass[twoside]{article}

\usepackage[accepted]{aistats2023}

\usepackage{packages}
\usepackage{multicol,wrapfig}
\usepackage{xcolor}
\usepackage{todonotes}

\newtheorem{lemma}{Lemma}

\usepackage[utf8]{inputenc} 
\usepackage[T1]{fontenc}    
\usepackage{url}            
\usepackage{booktabs}       
\usepackage{amsfonts}       
\usepackage{nicefrac}       
\usepackage{microtype}      
\usepackage{xcolor}         
\usepackage[inline]{enumitem}
\usepackage{multirow}

\def\ba{\mathbf{a}}
\def\bb{\mathbf{b}}
\def\bC{\mathbf{C}}
\def\bP{\mathbf{P}}

\def\bff{\mathbf{f}}
\def\bg{\mathbf{g}}
\def\bS{\mathbf{S}}
\def\bK{\mathbf{K}}
\def\bA{\mathbf{A}}
\def\bm{\mathbf{m}}
\def\bm{\mathbf{m}}
\def\bx{\mathbf{x}}
\def\by{\mathbf{y}}
\def\jacinput{\blacksquare}
\def\bSig{\mathbf{\Sigma}}
\usepackage{amsmath}%
\usepackage{MnSymbol}%
\usepackage{wasysym}%
 \usepackage[norule]{footmisc}

\begin{document}

\twocolumn[

\aistatstitle{Rethinking Initialization of the Sinkhorn Algorithm}

\aistatsauthor{ James Thornton \And Marco Cuturi }

\aistatsaddress{ University of Oxford\dagger \And  Apple}
]

\begin{abstract}
While the optimal transport (OT) problem was originally formulated as a linear program, the addition of entropic regularization has proven beneficial both computationally and statistically, for many applications. The Sinkhorn fixed-point algorithm is the most popular approach to solve this regularized problem, and, as a result, multiple attempts have been made to reduce its runtime using, e.g., annealing in the regularization parameter, momentum or acceleration.
The premise of this work is that \textit{initialization} of the Sinkhorn algorithm has received comparatively little attention, possibly due to two preconceptions: since the regularized OT problem is convex, it may not be worth crafting a good initialization, since \textit{any} is guaranteed to work; secondly, because the outputs of the Sinkhorn algorithm are often unrolled in end-to-end pipelines, a data-dependent initialization would bias Jacobian computations.
We challenge this conventional wisdom, and show that data-dependent initializers result in dramatic speed-ups, with no effect on differentiability as long as implicit differentiation is used. Our initializations rely on closed-forms for exact or approximate OT solutions that are known in the 1D, Gaussian or GMM settings. They can be used with minimal tuning, and result in consistent speed-ups for a wide variety of OT problems.
\end{abstract}

\section{Introduction} \label{sec:intro}
The optimal assignment problem and its generalization, the optimal transport (OT) problem, play an increasingly important role in modern machine learning. These problems define the Wasserstein geometry~\citep{SantambrogioBook,peyre2019computational}, which is routinely used as a loss function in imaging~\citep{schmitz2018wasserstein,janati2020multi}, but also used to reconstruct correspondences between datasets, as for instance in domain adaptation~\citep{courty2014domain,courty2017joint} or single-cell genomics~\citep{schiebinger2019optimal}. Several recent applications use OT to obtain an intermediate representation, as in self-supervised learning~\citep{caron2020unsupervised}, balanced attention~\citep{sander2022sinkformers}, parameterized matching~\citep{Sarlin_2020_CVPR}, differentiable sorting and ranking~\citep{adams2011ranking,cuturi2019differentiable,cuturi2020supervised,NEURIPS2020_ec24a54d}, differentiable resampling \citep{corenflos2021differentiable} and clustering~\citep{genevay2019differentiable}.

\textbf{Sinkhorn as a subroutine for OT.} A striking feature of all of the approaches outlined above is that they do not rely on the linear programming formulation of OT~\citep[\S9-11]{ahuja1988network}, but use instead an entropy regularized formulation~\citep{cuturi2013sinkhorn}. This formulation is typically solved with the ~\citeauthor{Sinkhorn67} algorithm~\citeyearpar{Sinkhorn67}, which has gained popularity for its versatility, efficiency and differentiability.

\textbf{Ever Faster Sinkhorn.} Given two discrete measures, the Sinkhorn algorithm runs a fixed-point iteration that outputs two optimal dual vectors, along with their objective--a proxy for their Wasserstein distance. Because Sinkhorn is often used as an inner routine within more complex architectures, its contribution to the total runtime may result in a substantial share of the entire computational burden. As a result, accelerating the Sinkhorn algorithm is crucial, and has been explored along two lines of works: through faster kernel matrix-vector multiplications, using geometric properties~\citep{2015-solomon-siggraph,altschuler2019massively,scetbon2020linear}, or by reducing the total number of iterations needed to converge, using e.g. an annealing regularization parameter~\citep{kosowsky1994invisible,schmitzer2019stabilized,xie2020fast}, momentum~\citep{thibault2021overrelaxed,lehmann2021note}, or \citeauthor{anderson1965iterative} acceleration \citeyearpar{anderson1965iterative}, as considered in~\citep{chizat2020faster}.

\textbf{Initialization as a Blind Spot.}
All methods above are, however, implemented by default by setting initial dual vectors naively at $\mathbf{0}$.

To our knowledge, initialization schemes have only been explored in a few restricted setups, such as semi-discrete settings in 2/3D~\citep{MEYRON201913}, or for discrete Wasserstein barycenter problems~\citep{cuturi2015smoothed}. We argue that careful initialization of dual potentials presents an overlooked opportunity for efficiency.

\textbf{Contributions.}
We propose multiple methods to initialize dual vectors. Contrary to concurrent and complementary work by~\citet{amos22}, our initializers are not trained, and not limited to fixed support setups. They require minimal hyperparameter tuning and result in small to negligible overheads. To do so, we leverage closed-form formulae and approximate solutions for simpler OT problems, resulting in the following procedures:
\begin{itemize}[leftmargin=.3cm,itemsep=.0cm,topsep=0cm]
\item We introduce a method to recover dual vectors when the primal problem solution is known in closed-form, and apply this to the non-regularized 1D problem. We show that initializing Sinkhorn with these vectors results in orders of magnitude speedups that can be readily applied to differentiable sorting and ranking.
\item When the ground cost is the squared L2 distance in $\mathbb{R}^d$, $d>1$, we leverage closed-form dual potential functions from the Gaussian approximation of source/target measures, and evaluate them on source points to initialize the Sinkhorn algorithm. We extend this by introducing an approximation of OT potentials for Gaussian \textit{mixtures}.
\item Finally we reformulate the multiscale approach of \citep[Alg. 3.6]{feydy2020geometric} as a subsample initializer.
\end{itemize}
We provide extensive empirical evaluation, and compare our approaches to other acceleration methods. We show that our initializations are robust and effective, outperforming existing alternatives, yet can also work in combination with them to achieve even better results.
\section{Background material on OT} \label{sec:background}
\subsection{Entropic Regularization and Sinkhorn} \label{sec:reg_ot}
Given two discrete probability measures $\mu=\sum_{i=1}^n a_i\delta_{x_i}$ and $\nu=\sum_{j=1}^{m} b_j\delta_{y_j}$ in $\mathcal{P}(\mathbb{R}^d)$, where ${\ba=(a_1,\dots, a_n)}$, ${\bb=(b_1,\dots, b_m)}$ are probability weights and ${(\bx_1,\dots, \bx_n)\in\mathbb{R}^{d\times n}}$, ${(\by_1,\dots, \by_m)\in\mathbb{R}^{d\times m}}$, the entropy regularized OT problem between $\mu$ and $\nu$ parameterized by $\varepsilon\geq 0$ and a cost function $c$ has two equivalent formulations, 
\begin{align}\label{eq:erot}
&\min_{\substack{\bP\in \mathbb{R}_+^{n\times m}, \bP\mathbf{1}_m=\ba, \bP^T\mathbf{1}_n=\bb}}\!\!\! \langle \mathbf{P}, \mathbf{C}\rangle - \varepsilon \langle\mathbf{P}, \log(\mathbf{P})-1\rangle\,, \\
&\!\!\!\!\!\!\max_{\bff\in\mathbb{R}^n, \bg\in\mathbb{R}^m}\!\!\!\! \mathcal{E}_{\mu,\nu,c,\varepsilon}(\bff, \bg)\!:=\!\langle\bff,\ba\rangle\! +\! \langle\bg,\bb\rangle -\!\varepsilon \langle e^{\tfrac{\bff}{\varepsilon}}, \mathbf{K} e^{\tfrac{\bg}{\varepsilon}}\rangle.\!\label{eq:erot_d}
\end{align}
where $\bC:=[c(\bx_i,\by_j)]_{i,j}$, with corresponding kernel $\mathbf{K} :=e^{-\bC/\varepsilon}$. While $(\bff,\bg)$ are unconstrained for $\varepsilon>0$, the regularization term converges as $\varepsilon\rightarrow 0$ to an indicator function that requires $\bff_i + \bg_j\leq c(\bx_i,\by_j)$.

    \begin{algorithm}[h]
    \caption{Sinkhorn's Algorithm}
    \label{algo:sinkhorn}
    \begin{algorithmic}[1]
        \STATE{{\bfseries Input:} $\ba,\bb, \mathbf{C}, \varepsilon>0, \omega>0, \bff^{(0)}, \bg^{(0)}$.}
        \STATE{\bfseries Initialize:} $\bff\leftarrow\bff^{(0)}, \bg\leftarrow\bg^{(0)}$
        \WHILE{not converged}
                \STATE $\bff \leftarrow \omega(\varepsilon\log \ba - \min_\varepsilon(\bC-\bff\oplus\bg)) + \bff$
                \STATE $\bg \leftarrow \omega(\varepsilon\log \bb - \min_\varepsilon(\bC^T-\bg\oplus\bff)) + \bg$
        \ENDWHILE
        \STATE{{\bfseries Return} {$\bff, \bg$ }}
    \end{algorithmic}
    \end{algorithm}
%

\textbf{The Sinkhorn Algorithm.} \Cref{algo:sinkhorn} describes a sequence of updates to optimize $\bff,\bg$ in~\eqref{eq:erot_d}. When $\omega=1$, these updates correspond to cancelling alternatively the gradients $\nabla_{1} \mathcal{E}_{\mu,\nu,c,\varepsilon}(\bff, \bg)$ (line 4) and $\nabla_{2} \mathcal{E}_{\mu,\nu,c,\varepsilon}(\bff, \bg)$ (line 5) of the objective in~\eqref{eq:erot_d}. These updates use the row-wise soft-min operator $\min_\varepsilon$, defined as:
$$ \text{Given } \bS=[\bS_{i,j}],\,\text{min}_\varepsilon(\bS) := [-\varepsilon \log\left( \mathbf{1}^Te^{-\bS_{i,\cdot}/\varepsilon}\right)]_i\,,$$
and the tensor addition notation $\bff\oplus\bg = [\bff_i+\bg_j]_{i,j}$. The runtime of the Sinkhorn algorithm hinges on several factors, notably the choice of $\varepsilon$.
Several works report that hundreds of iterations are typically required when using fairly small regularization $\varepsilon$ (e.g. 500 in~\citealt[App.B]{salimans2018improving}).
These scalability issues are compounded in advanced applications whereby multiple Sinkhorn layers are embedded in a single computation or batched across examples~\citep{cuturi2019differentiable,NEURIPS2020_ec24a54d,cuturi2020supervised}.
 To mitigate runtime issues, popular acceleration techniques such as fixed~\citep{thibault2021overrelaxed} or adaptive~\citep{lehmann2021note} momentum approaches,
 as well as Anderson acceleration~\citep{chizat2020faster} have been considered.
While acceleration methods are known to work well when initialized not too far away from optima~\citep{d2021acceleration}, all common implementations~\citep{flamary2021pot,cuturi2022optimal} initialize these vectors to $(\mathbf{0}_n,\mathbf{0}_m)$.

\subsection{Dual Variables in the Sinkhorn Algorithm} \label{sec:ot}
\textbf{On starting closer to the solution.}
While the Sinkhorn algorithm will converge with any initialization, the speed of convergence is bounded by ~\citep[Rem. 4.14]{peyre2019computational}:
\begin{align}\label{eq:bound}
\|\bff^{(\ell)} - \bff^\star\|_{\text{var}} \leq \|\bff^{(0)} - \bff^\star\|_{\text{var}}\lambda(\bK)^{2\ell}\,,
\end{align}
where $\bff^{(\ell)}$ denotes the potential vector $\bff$ obtained after running ~\Cref{algo:sinkhorn} for $\ell$ iterations, $\bff^\star$ the optimal potential, and deviation is measured using the variation norm. $\lambda(\bK)$ reflects conditioning in $\bK$~\citep[Theorem 4.1]{peyre2019computational}, determined by the range and magnitude of costs evaluated on $(\bx_i,\by_j)$ pairs relative to $\varepsilon$. Since $0<\lambda(\bK)<1$, the Sinkhorn algorithm converges more slowly as $\lambda(\bK)$ approaches $1$. The motivation to obtain a better initialization relies on targeting the initial gap in $\|\bff^{(0)} - \bff^\star\|_{\text{var}}$.

\textbf{Two or One Dual Initializations?}
While~\Cref{algo:sinkhorn} lists two initial vectors $(\bff^{(0)}, \bg^{(0)})$, a closer inspection of the updates shows that only a single dual variable is needed: when starting with an iteration updating $\bg$, only $\bff^{(0)}$ is required (the reference to $\bg$ is only there for numerical stability). Conversely, only $\bg^{(0)}$ is required when updating $\bff$. Since only one is needed, we supply by default the smallest vector when $n\ne m$, and set the other to $\mathbf{0}$.

\textbf{Differentiability and Dual initialization.} 
Any output of the Sinkhorn fixed-point algorithm can be differentiated using unrolling~\citep{adams2011ranking,hashimoto2016learning,genevay2018learning,genevay2019differentiable,cuturi2019differentiable,caron2020unsupervised}. This approach has, however, two drawbacks: its memory footprint grows as $L(n+m)$, where $L$ is the number of iterations needed to converge, and, more fundamentally, it prevents us from using more efficient steps, such as adaptive momentum and acceleration, because they typically involve non-differentiable operations. These issues can be avoided by relying instead on implicit differentiation~\citep{luise2018differential,cuturi2020supervised,xie2020fast,cuturi2022optimal}, which only requires access to solutions $\bff^\star, \bg^\star$ to work. We recall how this can be implemented for completeness. Introducing the following notations: 
$$
\begin{aligned}
F:& \,\mu,\nu, c, \varepsilon \mapsto \bff^\star, \bg^\star, \text{optimal solutions to~\eqref{eq:erot_d}}\,,\\
H:& \,\mu,\nu, c, \varepsilon,\bff ,\bg \mapsto \begin{bmatrix}\nabla_1\mathcal{E}_{\mu,\nu,c,\varepsilon}(\bff,\bg)\\\nabla_2\mathcal{E}_{\mu,\nu,c,\varepsilon}(\bff,\bg)\end{bmatrix}\,,
\end{aligned}
$$
one has that $H(\mu,\nu, c, \varepsilon,F(\mu,\nu, c, \varepsilon))=\mathbf{0}_{n+m}$, which is the root equation that can be used to instantiate the implicit function theorem, to recover the Jacobian of the outputs of $F$ (i.e. $\bff^\star ,\bg^\star$) w.r.t. \textit{any} variable ``$\jacinput$'' within inputs. As a result, the transpose-Jacobian of $F$ applied to any perturbation of the size of $\jacinput$ (the only operation needed to implement reverse-mode differentiation) is recovered as (where $\dots$ is a shorthand notation for $\jacinput,(\bff^\star,\bg^\star)$):
$$J_{F,\jacinput}(\dots)^Tz =- J_{H,\jacinput}(\dots)^T
 \left(J_{H,(\bff,\bg)}(\dots)^{T}\right)^{-1}\mathbf{z} $$
All of these operations can be instantiated easily using \texttt{vjp} Jacobian operators~\citep{bradbury2018jax} and linear systems that rely on linear functions (rather than matrices) as detailed in~\citep{cuturi2022optimal}. These computations only require access to optimal values $\bff^\star, \bg^\star$, not the computational graph that was needed to reach them.

\subsection{Closed-Form Expressions in Optimal Transport}\label{sec:back_closed}
A few closed-forms for \textit{unregularized} ($\varepsilon=0$) OT are known.
Some of these closed forms rely on the \citeauthor{Monge1781} formulation of OT, recalled for completeness for two measures $\mu, \nu \in\mathcal{P}(\mathbb{R}^d)$ in~\eqref{eq:monge}, using the push-forward $\sharp$ notation, as well as the dual formulation of OT in~\eqref{eq:ot_dual}, using the convention $f^c(\by):=\min_{\bx} c(\bx,\by) - f(\bx)$, the $c$-transform of $f$.
\begin{align}
\min_{\substack{T:\mathbb{R}^d\rightarrow\mathbb{R}^d\\ T_\sharp \mu = \nu}} &\int\!\! c(\bx,T(\bx))\mathrm{d}\mu(\bx) .\label{eq:monge}\\
\max_{f:\mathbb{R}^d\rightarrow \mathbb{R}} &\int\!\! f \mathrm{d}\mu+\!\!\int\!\! f^c \mathrm{d}\nu\,. \label{eq:ot_dual}
\end{align}
We review two relevant cases, where either an optimal coupling $\bP^\star$ (for $\varepsilon=0$) in the primal formulation of \eqref{eq:erot}, or an optimal map $T^\star$ to~\eqref{eq:monge} can be obtained in closed form. We show in \S\ref{sec:methods} how these solutions can be leveraged to recover initialization vectors $\bff^{(0)}$ and $\bg^{(0)}$ for Alg.~\ref{algo:sinkhorn}. 

\textbf{OT in 1D.}
For univariate data ($d=1)$, and when the cost function $c$ is such that $-c$ is supermodular $(\partial c/\partial x \partial y<0)$, a solution $\bP^\star$ to~\eqref{eq:erot} can be recovered in closed form~\citep[\S3]{chiappori2017multi,SantambrogioBook}. Writing $\sigma$, $\rho$ for sorting permutations of the supports of $\mu$ and $\nu$, $x_{\sigma_1} \leq \dots \leq x_{\sigma_n}$ and $y_{\rho_1}\leq \dots\ y_{\rho_m}$, a solution $\bP^{\star}$ is given by the \textit{north-west corner} solution $\textbf{NW}(\ba_\sigma,\bb_\rho)$, where $\ba_\sigma$ and $\bb_\rho$ are the weight vectors $\ba,\bb$ permuted using $\sigma$ and $\rho$ respectively~\citep[\S3.4.2]{peyre2019computational}. 

\begin{figure}
\centering
\includegraphics[width=\linewidth]{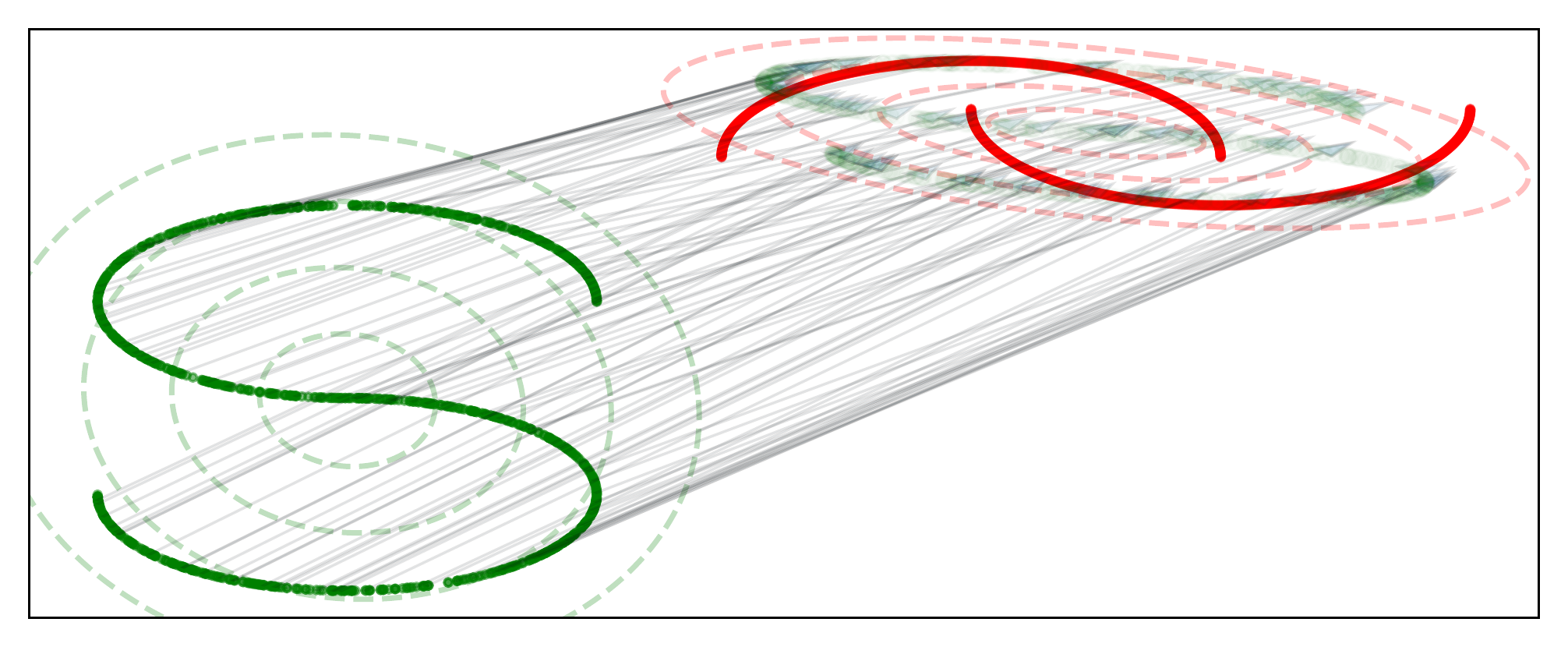}
\caption{Transport map (black) from Gaussian approximations (dashed) of S-curve (green) and two-moons (red)}
\label{fig:2d_toy}
\vspace{0.1cm}
\end{figure}

\textbf{Gaussian.} The \citeauthor{Monge1781} formulation of the OT problem~\eqref{eq:monge} from a Gaussian measure $\mathfrak{N}_1 =\mathcal{N}(\bm_1,\bSig_1), \bSig_1>0$, to another $\mathfrak{N}_2= \mathcal{N}(\bm_2, \bSig_2)$, is solved by (see also Fig.~\ref{fig:2d_toy}):
$$T^\star(\bx):= \mathbf{A}(\bx-\bm_1) + \bm_2, \mathbf{A}=\bSig^{-\frac{1}{2}}_1(\bSig^{\frac{1}{2}}_1\bSig_2\bSig^{\frac{1}{2}}_1)^\frac{1}{2}\bSig^{-\frac{1}{2}}_1.$$ The optimal dual \textit{potential} $f^\star$ is a quadratic form given by 
\begin{equation}\label{eq:G-potential}
f^\star(\bx)=\tfrac{1}{2}\bx^T(\mathbf{I}-\mathbf{A})\bx + (\bm_2-\mathbf{A} \bm_1)^T\bx\,,
\end{equation}
which recovers $T^\star=\text{Id}-\nabla f^\star$.
The OT cost between $\mathfrak{N}_1$ and $\mathfrak{N}_2$ is known as the Bures-Wasserstein distance:
\begin{equation}\label{eq:BW}
\begin{aligned}
W_2^2(\mathfrak{N}_1, \mathfrak{N}_2)&=\|\bm_1 - \bm_2\|^2 + \mathcal{B}^2_2(\Sigma_1, \Sigma_2)\,,\\
\mathcal{B}^2_2(\Sigma_1, \Sigma_2) :&= \text{tr}(\bSig_1 + \bSig_2 - 2 (\bSig^{\frac{1}{2}}_1\bSig_2\bSig^{\frac{1}{2}}_1)^\frac{1}{2})\,.
\end{aligned}
\end{equation}

\section{Crafting Sinkhorn Initializations}\label{sec:methods}
We present important scenarios where careful initialization can dramatically speed up the Sinkhorn algorithm. We start with the 1D case (\S\ref{subsec:sorting}), where entropic transport has been used recently as a possible approach to obtain differentiable rank and sorting operators. We follow with the generic and by now standard multivariate OT problem in $\mathbb{R}^d$ with squared-L2 ground cost, using Gaussian approximations (\S\ref{subsec:gauss_init}) and an extension to mixtures (\S\ref{subsec:gmms}). 

\subsection{Initialization for 1D Regularized OT}\label{subsec:sorting}
\textbf{Ranking as an OT problem.}
Using a cost $c$ on $\mathbb{R}\times \mathbb{R}$ such that $\partial c/\partial x \partial y<0$, 
sorting the entries of a vector $\mathbf{x} = (x_1,\dots, x_n) \in \mathbb{R}^n$ can be recovered using a solution $\bP^\star$ to ~\eqref{eq:erot}, setting $\varepsilon=0$, $\ba=\mathbf{1}_n/n$, and $\nu$ to a uniform measure on $n$ increasing numbers, e.g. $\mathbf{y}= (1,2,\ldots,n)$. The ranks of the entries of $\mathbf{x}$ are then $n\bP^\star \mathbf{z}$, where $\mathbf{z}=(1,2,...,n)$, and its sorted entries as $n\bP^{\star T} \mathbf{x}$~\citep{cuturi2019differentiable}.

\textbf{Differentiable Ranking.} A differentiable and fractional soft sorting/ranking operator can be derived from entropy regularized couplings, using instead a solution $\bP_\varepsilon$ to (\ref{eq:erot}, $\varepsilon>0$) to form $n\bP_\varepsilon \mathbf{z}$ and $n\bP_\varepsilon^T \bx$~\citep{cuturi2019differentiable}, with the possibility to use a different target size $m$ or non-uniform weights $\ba,\bb$. A practical challenge of that approach is that the number of Sinkhorn iterations needed for the coupling to converge can be typically quite large, see ~\Cref{fig:comp_base} and further results in ~\Cref{sec:app_softsort}.

\textbf{Dual 1D Initializers.} Regularized 1D OT problems often require a small regularization $\varepsilon$ to be meaningful, in order to recover rank approximations that are not too smoothed, which then requires many Sinkhorn iterations to converge. To address this, we introduce an initializer using potentials for the non-regularized problem ($\varepsilon=0$). Our strategy to pick initialization vectors for ~\Cref{algo:sinkhorn} is upon first glance deceptively simple: sort $\bx$, recover a primal solution $\bP^\star$ (the North-West corner solution) that is guaranteed to solve~\eqref{eq:erot}; turn it into a pair of optimal dual vectors $\bff^\star_0,\bg^{\star}_0$ for the same unregularized problem, and seed them to Alg.~\ref{algo:sinkhorn} to solve~\ref{eq:erot_d} with $\varepsilon>0$.
While obtaining $\bP^\star$ only requires a sort, efficiently recovering a corresponding dual pair ($\bff^\star_0,\bg^\star_0$) is less straightforward. In principle, duals may be obtained by solving an elementary cascading linear system using primal-dual conditions~\citep[\S3.5.1]{peyre2019computational}. That approach does not always work, however, when the size of the support of $\bP^\star$ is strictly smaller than $n+m-1$ (it results in a system that has less equalities than variables), which is the case in the original ranking problem, where $n=m$. 
\citet[Alg.1]{pmlr-v151-sejourne22a} propose an algorithm to construct $\bff^\star_0,\bg^{\star}_0$ in $n+m$ sequential operations, interlaced with conditional statements. We consider a more generic algorithm that works in higher dimensions, but which, when particularized to the 1D case, results in the \textsc{DualSort} Algorithm~\ref{algo:sorting_init} (see also~\Cref{sec:dualsort_conv}), a parallel approach with larger $\mathcal{O}(nm)$ complexity, but simpler to deploy on GPU, since it only requires a handful of iterations to converge, each directly comparable to that of the Sinkhorn algorithm. See application to experiments in \S\ref{subsec:diffsort} and \S\ref{subsec:softerr}.

\begin{algorithm}[h]
\caption{\textsc{DualSort} Initializer}\label{algo:sorting_init}
        \begin{algorithmic}[1]
            \STATE{{\bfseries Input:} { Cost matrix $\mathbf{C}=[c(x_{\sigma_i},y_{\rho_j})]$ for the sorted entries of input vectors $\bx, \by$ entries, see \S\ref{sec:back_closed}.}}
            \STATE{\bfseries Initialize:} $\mathbf{f} = 0$
            \WHILE{not converged}
            \STATE{$\bff \gets \min_{\text{axis=1}} \left(\mathbf{C}-\textrm{diag}(\mathbf{C})\mathbf{1}^T+\bff\mathbf{1}^T\right)$}
  
            \ENDWHILE
            \STATE{{\bfseries Return} {$\mathbf{f}$}}
\end{algorithmic}
\end{algorithm}

\subsection{Computing Dual Initializers from Gaussian OT}\label{subsec:gauss_init}

\looseness -1
\textbf{From optimal potentials to dual initializers.}
We leverage Gaussian approximations to obtain an efficient initializer, coined \textsc{Gaus}, for the Sinkhorn problem, when $c(x,y)=\|x-y\|^2_2$, notably when $n\gg d$. 

To do so, and given two discrete empirical measures $\mu$  
and $\nu$, 
compute their empirical means and covariance matrices $(\bm_\mu, \mathbf{\Sigma}_\mu)$ and $(\bm_\nu, \mathbf{\Sigma}_\nu)$, to recover a dual potential \textit{function} $f^\star$ from ~\eqref{eq:G-potential} that solves the Gaussian dual OT problem, where $\bA$ in that equation can be obtained by replacing $\bSig_1$ with $\mathbf{\Sigma}_\mu$ and $\bSig_2$ with $\mathbf{\Sigma}_\nu$. Next, evaluate that quadratic potential on all observed points of the first measure $[\bff^{(0)}]_i \gets f^\star(\bx_i)$ (or alternatively the second measure if $m<n$) to seed the Sinkhorn algorithm.
\begin{table}[ht]
\centering
\caption{Toy examples, $n=m=1024$, $d=2$, $200$ runs.}
\label{tab:gaus_toy}
\begin{tabular}{lll}
    \toprule
    \multirow{0}{0pt}{Dataset}  &  \multicolumn{2}{c}{\# Iterations $(\textrm{mean} \pm \textrm{std})$ \hfill} \\ 
         & Init $\mathbf{0}$ & Init Gaus    \\
    \midrule
    2-moons &   $120.0 \pm 0.0$ & $\mathbf{11.0 \pm 0.0}$    \\
    S curve / 2-moons &  $137.2 \pm 16.7$  &    $\mathbf{49.6 \pm 14.8}$     \\
    3 Gaussian blobs  &  $236.0 \pm 24.3$  & $\mathbf{45.4 \pm 9.7}$  \\
    \bottomrule
\end{tabular}
\end{table}

\textbf{Complexity.}
Solving OT on the Gaussian approximations of $\mu, \nu$, requires computing means and covariance matrices $\mathcal{O}((n+m)d^2)$, as well as matrix square-roots and their inverse, using the Newton-Schulz iterations~\citep{higham2008functions} at cost $\mathcal{O}(d^3)$. The  \textsc{Gaus} initializer is therefore particularly relevant in settings where $d \ll n$, which is typically the regime where OT has found practical relevance.

\textbf{Implementation.} 
Our experiments show that \textsc{Gaus} often works significantly better than the default null initialization, notably with toy datasets (see Table~
\ref{tab:gaus_toy}), but also when computing OT on latent space embeddings as shown in \S\ref{sec:clustering} and \S\ref{sec:dpf}, or to word-embeddings as demonstrated in \S\ref{sec:word_embeddings}. The overhead induced by the computations of dual solutions is naturally dictated by the tradeoff between $n$ (the number of points) and $d$ (their dimension). In all cases considered here that overhead is negligible, but explored with more care in Appendix~\ref{sec:overhead}. Note that many of the matrix-squared-roots computations can be pre-stored for efficiency, if the same measure $\mu$ is to be compared repeatedly to other measures.

\subsection{Gaussian Mixture Approximations}\label{subsec:gmms}
The Gaussian initialization approach can be extended to Gaussian mixture models (GMMs), resulting in greater flexibility, yet pending further approximations.
This requires the additional cost of pre-estimating GMMs for each input measure. By \textit{further} approximations above, we refer more explicitly to the fact that, unlike for single Gaussians, we do not have access to closed-form OT solutions between GMMs, but instead only ``efficient'' couplings that return a cost that is an upper-bound on the true Wasserstein distance between two GMMs, as introduced next.

\textbf{OT in the space of Gaussian measures.}
Given two Gaussian mixtures $\rho=\sum_{k=1}^{K}\alpha_k\rho_k$ and $\tau=\sum_{k=1}^{K}\beta_k\tau_k$, assuming each $\rho_k$ and $\tau_k$ is itself a Gaussian measure, and that weights $\alpha_k$ and $\beta_k$ sum to $1$. It was proposed in \citep{chen2018optimal} to approximate the continuous OT problem between $\rho$ and $\tau$ in the space $\mathbb{R}^d$ as a discrete OT problem in the space of mixtures of Gaussians, where each mixture is a discrete measure on $K$ atoms (each atom being a Gaussian), and the ground cost between them is set to the pairwise Bures-Wasserstein distance, forming a cost matrix for~\eqref{eq:erot} as $\mathbf{C}=[W_2^2(\rho_i,\tau_j)]_{ij}$ using \eqref{eq:BW}. That optimization results in two potentials $\tilde{\bff}$ and $\tilde{\bg}\in\mathbb{R}^K$ that solve the corresponding regularized $K\times K$ OT problem.

\begin{figure}[h]
    \centering
    \includegraphics[width=\linewidth]{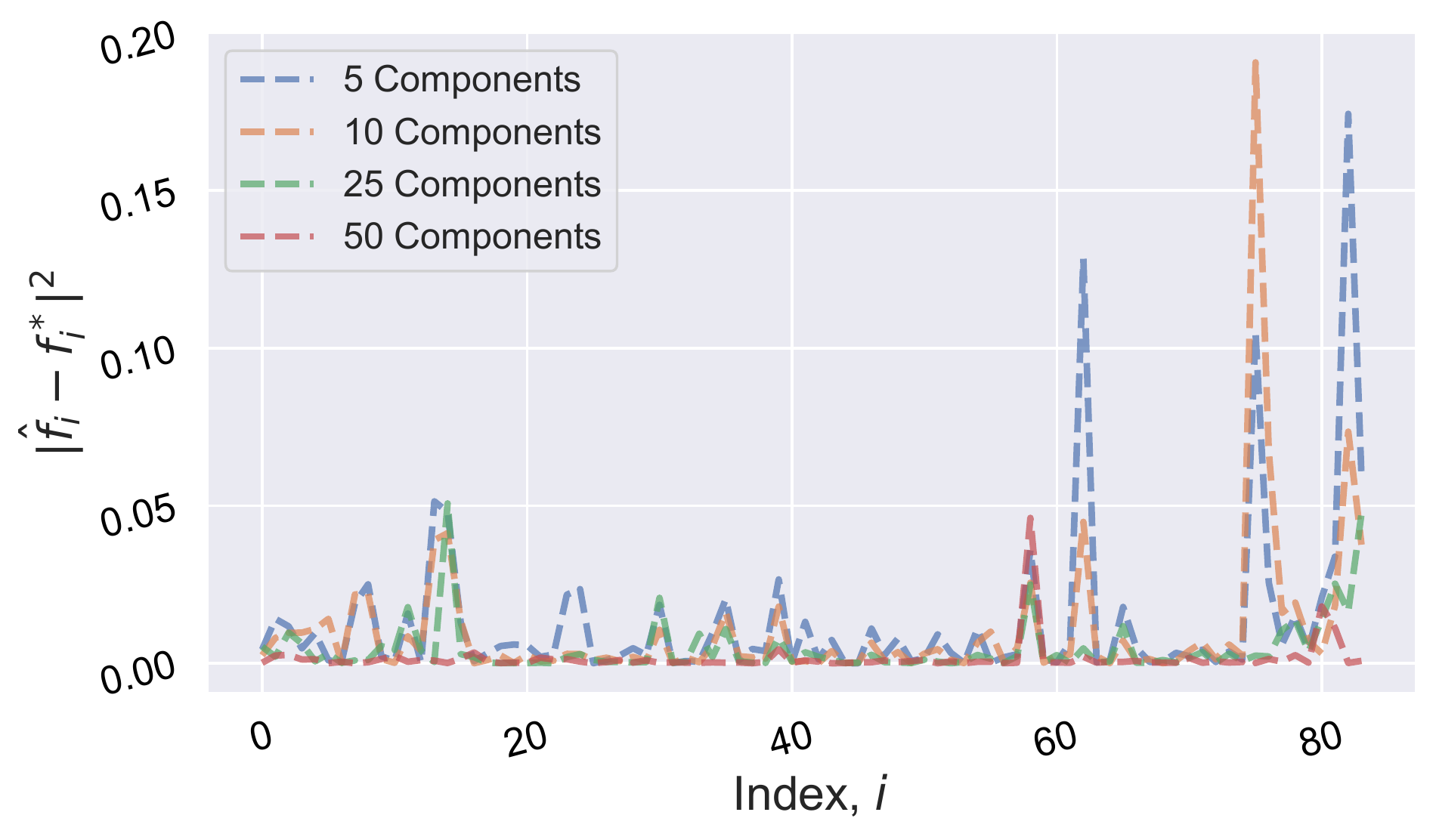}
    \caption{Gap between the true dual $\bff^\star$ and the GMM approximate dual, for a pair of measures of word embeddings, as a function of $K$, the number of mixture components.}
    \label{fig:gmm_approx}
\end{figure} 

\textbf{Approximating Dual Potentials with GMMs.}
Our proposed initializer, \textsc{GMM}, is computed as follows. Given two empirical measures $\mu,\nu$, we fit first two $K$-component GMMs $\tau$ and $\rho$, then obtain two potential vectors $\tilde{\bff},\tilde{\bg}\in\mathbf{R}^K$ using the Sinkhorn algorithm on a $K\times K$ problem, as described above.
From those potentials, we propose to compute an approximate $\hat{f}$ dual potential function:
\begin{equation}\label{eq:gmm_approx} 
\hat{f}(\bx)= \tilde{\bff}^T p(\bx),\, [\,p(\bx)]_k =  \frac{\alpha_k d\rho_k(\bx)}{\sum_{l=1}^K \alpha_l d\rho_l(\bx)}\,.
\end{equation} 
that is then evaluated on all $n$ points of $\mu$. Intuitively this approximation interpolates continuously the $K$ potentials depending on probability within mixture. This recovers, in the limit where $K\rightarrow n$, $n$ components with means $(x_i)_i$ and zero covariance, resulting in the original potential $\bff^{\star}$.

\textbf{Complexity.} Fitting GMMs cost $\mathcal{O}(nKd^2)$. Computing the Bures-Wasserstein distances between two Gaussian measures would have complexity $\mathcal{O}(d^3)$ for full covariance matrices and $\mathcal{O}(d)$ for diagonal. Computing the cost matrix for the GMM OT problem would then amount to $\mathcal{O}(K^2d^3)$ or $\mathcal{O}(K^2d)$. Since naive Sinkhorn requires $\mathcal{O}(Ln^2)$ to run between pointclouds of size $n$ for $L$ iterations, and so the proposed GMM initialization may provide, very roughly and not taking into account pre-storage, efficiency gains when $K^2d \ll n^2$.

\subsection{Initialization via Subsampling}

We next bring attention to a multi-scale approach described in detail in \citep[Alg. 3.6]{feydy2020geometric}, which is a competitive baseline for comparison. Although not how originally described, this approach may be framed as a Sinkhorn initializer which we call the \textsc{Subsample} initializer. The \textsc{Subsample} initializer builds on the idea of the out-of-sample extrapolated entropic potentials~\citep{pooladian} that are derived readily from a first resolution of the OT problem on a subset of points. Let $\breve\mu$, $\breve\nu$ denote uniformly subsampled measures of $\mu$ and $\nu$ of size $\breve{n}\ll n$ and $\breve{m}\ll m$. Write $\breve{\mu}=\tfrac{1}{\breve{n}}\sum_i \delta_{w_i}$, $\breve{\nu}=\tfrac{1}{\breve{m}}\sum_i \delta_{z_i}$ and write $\breve{\bff},\breve{\bg}$ the optimal vector dual potentials obtained for~\eqref{eq:erot_d} for the same regularization $\varepsilon$ and cost, but using $\breve{\mu}$ and $\breve{\nu}$ instead. An initializer for $\bff^{(0)}$,  can be then defined by using the entropic potential function derived from $\breve{\bg}$ (or, alternatively from $\breve{\bff}$ if $n\ll m$):
\begin{equation}\label{eq:entropicmap}
[\bff^{(0)}]_i = \breve{f}(\bx_i), \text{with}\,\breve{f}:\bx \mapsto -\varepsilon \log \tfrac{1}{\breve{m}}\!\sum_{j=1}^{\breve{m}} e^{\frac{\breve{\bg}_j - c(\bx_, \mathbf{z}_j)}{\varepsilon}}.
\end{equation}
Although more general than the \textsc{GMM} initializer, the \textsc{Subsample} initializer requires running Sinkhorn on a subsample of points $\breve{n},\breve{m}$ that is typically larger than the $K\times K$ problem induced by $K$-components GMMs. While this may show in runtime costs, as in \Cref{fig:word_emb_timing}, the Sinkhorn initializer, on the other hand, not affected by large dimensions.

\section{Experiments} \label{sec:experiments}
In this section we illustrate the benefits of our proposed initialization strategies. In particular, we apply \textsc{DualSort} for differentiable sorting and soft-$0/1$ loss from \citep{cuturi2019differentiable}. We investigate Gaussian (\textsc{Gaus}) initializers for deep differentiable clustering from \citep{genevay2019differentiable} and differentiable particle filtering from \citep{corenflos2021differentiable}. Finally, we showcase \textsc{GMM} initializers with a document similarity task. The purpose of these experiments is to show the benefit of the initializer and not the performance in the particular task, or in claiming these tasks are original. With that in mind, we have not performed extensive network parameter tuning, though we do include some performance metrics to illustrate that the setups are reasonable. Further experimental details are given in Appendix~\ref{sec:app_exp}. Experiments were carried out using \texttt{OTT-JAX} \citep{cuturi2022optimal}, notably acceleration methods for comparison, but also, when relevant, implicit differentiation of Sinkhorn's outputs.

We compare our proposed approaches to the default $\mathbf{0}$ initialization typical in most Sinkhorn implementations, in addition to fixed ~\citep{thibault2021overrelaxed} and adaptive \citep{lehmann2021note} momentum,   $\varepsilon-decay$,
 as well as Anderson acceleration~\citep{chizat2020faster}.

\subsection{Differentiable Sorting}\label{subsec:diffsort}
Arrays of size ${n \in \{16, 32, 64, 128, 256, 512, 1024\}}$ were sampled in this experiment from the Gaussian blob dataset \citep{scikit_learn} for $200$ different seeds. For each seed, $1$-dimensional Gaussian data was generated from $5$ random centers with centers uniformly distributed in $(-10, 10)$ with standard deviation $3$. The Sinkhorn algorithm was then ran with the proposed initialization, \textsc{DualSort}, and with the default zero initializer, labelled $\mathbf{0}$. Other Sinkhorn acceleration methods were also investigated including Anderson acceleration (And$=5$), momentum ($\text{mom.}=1.05$), regularization decay ($\varepsilon \textrm{ decay} = 0.8$) and adaptive momentum (adapt$=10$, meaning adaptation is recomputed after 10 iterations). The parameter values for these competing methods were pre-tuned following an initial hyper-parameter sweep.
      
\begin{figure}[h]
         \centering
         \includegraphics[width=\linewidth]{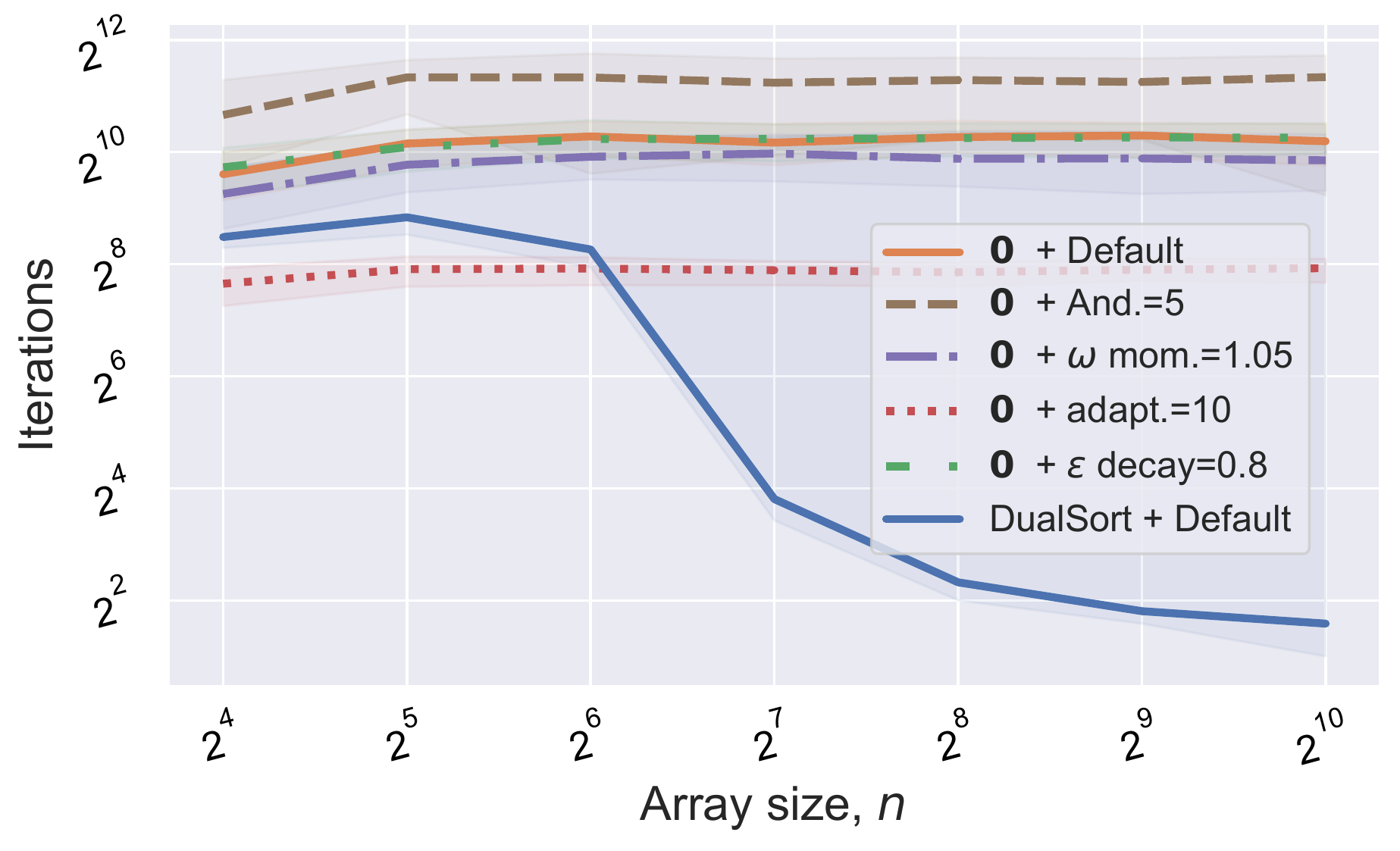}
         
         \caption{\textsc{DualSort} with a default Sinkhorn setup dominates all existing acceleration methods implemented when run with a default $\mathbf{0}$ initialization. We plot median, upper and lower quartiles of iterations needed to converge over $200$ seeds for various array sizes (iterations for \textsc{DualSort} include steps for the primal-dual procedure).}
         \label{fig:comp_base}
\end{figure}

\begin{figure}[h]
        \centering
         \includegraphics[width=\linewidth]{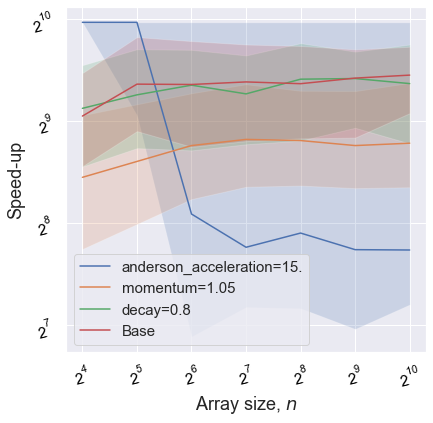}
         \caption{Relative speed-up (higher is better).  Median, upper and lower quartiles of iterations needed to converge over $200$ seeds for various array sizes (iterations for \textsc{DualSort} include steps for the primal-dual procedure).}
         \label{fig:enhance_comp}
\end{figure}

Figures \ref{fig:comp_base} and \ref{fig:enhance_comp} illustrate the dramatic speed-up effect from using the \textsc{DualSort} procedure, with just $3$ vectorized iterations. Figure \ref{fig:comp_base} compares Sinkhorn algorithm with initialization to Sinkhorn enhanced through other acceleration method. Figure \ref{fig:enhance_comp} illustrates the relative-speed up from including initialization along with other enhancements where speed-up is defined as the ratio of iterations using the zero initializer and the \textsc{DualSort} initializer, hence $>1$ indicates an improvement using \textsc{DualSort}. \textsc{DualSort} complements existing acceleration methods. When the \textsc{DualSort} initializer is paired with other acceleration methods, we still observe, no matter which one is used, very large speedups.

\textbf{Runtime cost.} The \textsc{DualSort} initializer's runtime cost is negligible and took just $0.0012$ seconds (s) to run for all experiments. The resulting absolute speed-up was $0.06s$ to $0.13$s per OT problem. See further timing details in Appendices \ref{sec:app_softsort}. Note that this speed-up is compounded when running many thousands of OT problems.

\begin{table}[ht]
\caption{Average time in seconds for DualSort with 3 iterations and Sinkhorn iterations to convergence over 200 soft sorting problems for Gaussian blob data of dimension $n$}
\centering
{\small{
\begin{tabular}{llcc}
\centering
\textbf{$n$} & \textbf{Initializer} & \textbf{Initialization}                              & \textbf{Iterations} \\
 \midrule
$32$         & $\mathbf{0}$         &     -                                                 & $0.28$                 \\
             & DualSort             & $0.0012$ & $0.22$                 \\
              \midrule
$64$         & $\mathbf{0}$         &     -                                                 & $0.22$                 \\
             & DualSort             & $0.0012$                                             & $0.088$                 \\
              \midrule
$128$        & $\mathbf{0}$         &         -                                             & $0.17$                 \\
             & DualSort             & $0.0012$                                             & $0.066$                 \\
              \midrule
$256$        & $\mathbf{0}$         &     -                                                 & $0.17$                 \\
             & DualSort             & $0.0012$                                             & $0.049$                 \\
              \midrule
$512$        & $\mathbf{0}$         &      -                                                & $0.13$                 \\
             & DualSort             & $0.0012$                                              & $0.050$                \\
              \midrule
$1024$       & $\mathbf{0}$         &       -                                               & $0.14$                 \\
             & DualSort             & $0.0012$                                              & $0.058$            \\
             \bottomrule
\end{tabular}}}
\end{table}

\subsection{Soft Error Classification} \label{subsec:softerr}
    The following experiment demonstrates the differentiability of the soft-sorting and ranking operations as well as how the \textsc{DualSort} initializer improves computational performance for real tasks. 
        \begin{figure}[h]
            \centering
            \includegraphics[width=\linewidth]{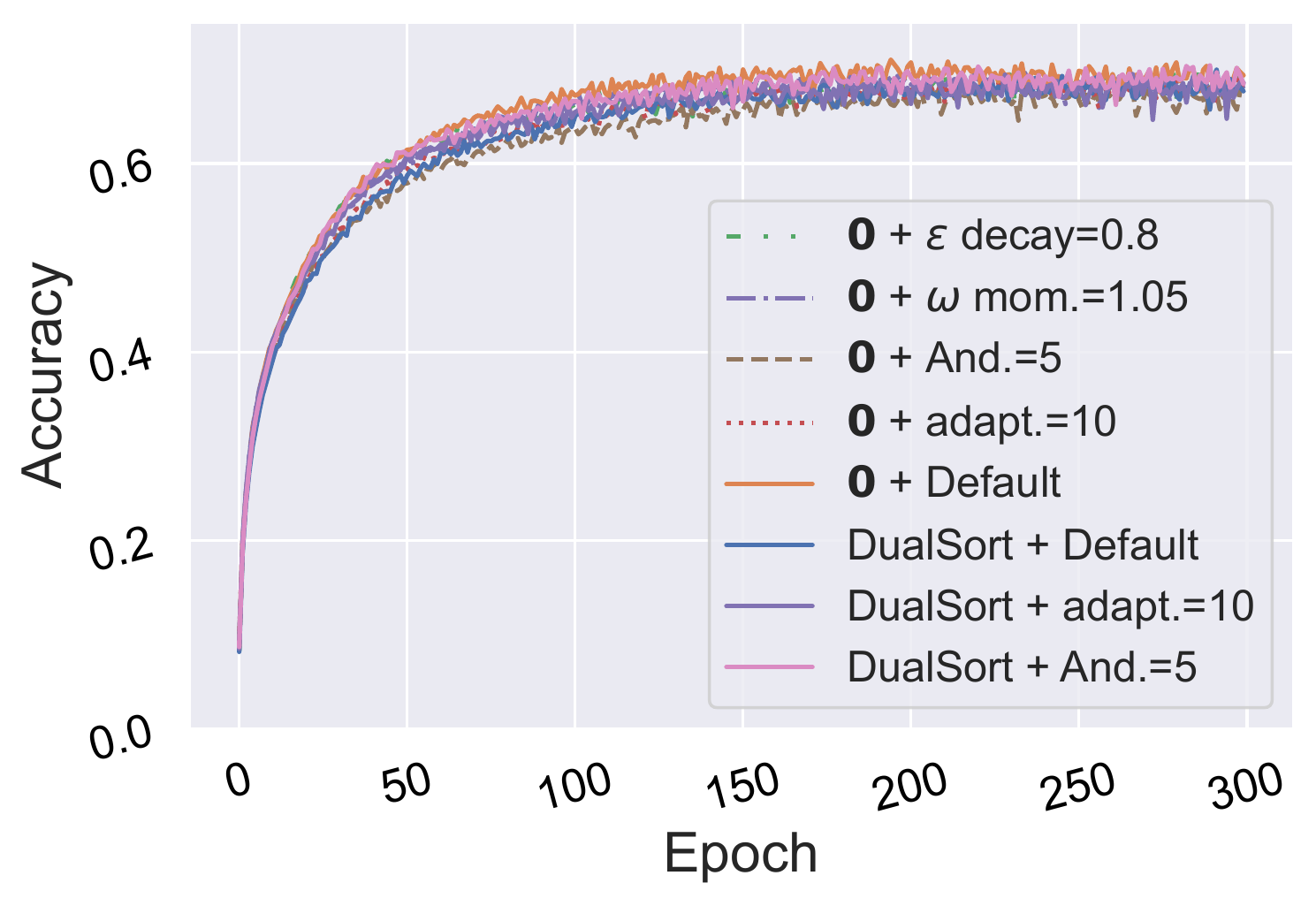}
        \caption{Accuracy of CNN classifier by Sinkhorn methods for CIFAR-100 with soft-error loss and $\varepsilon=0.01$}
        \label{fig:cifar_classifier}
    \end{figure}
    Let ${h_\theta: \mathcal{X} \rightarrow \mathbb{R}^K}$ be a parameterized $K$-label classifier and $R$ the differentiable ranking operator described in \S\ref{subsec:sorting}. For input $x\in\mathcal{X}$, the soft-$0/1$ loss (or soft-error) evaluated at labeled $(x,y)$, $y\leq K$, is therefore $\max(0,K-R(h_\theta(x))_y)$, see \citep{cuturi2019differentiable} for details.
        
    \begin{table}
    \centering
    {\small{
    \caption{Soft-Error: CIFAR 100, mean $\pm$ std of Sinkhorn iter./ training step, over $10$ seeds}
    \label{tab:soft_error}
    \centering
    \begin{tabular}{lll}
    \toprule
    {}               & {Iterations}  & Runtime ($\times 10^{-2}$s) \\ \midrule
    {Zero}                      & $ 17.9 \pm 0.1$  & $ 8.23 \pm 0.2$     \\
    {Anderson}                  & $ 12.3 \pm 0.2$  & $ 5.74 \pm 0.2 $     \\
    {Momentum}                  & $ 15.7 \pm 0.2$  & $ 7.70 \pm 0.2$     \\
    {Adaptive}                  & $ 15.2 \pm 0.2$  & $ 7.38 \pm 0.3$     \\
    {$\varepsilon$-decay}       & $ 17.0 \pm 0.1$  & $  7.99\pm 0.2$     \\
    {\textsc{DualSort}}                  & ${ 9.7 \pm 0.1}$ & $ 5.07 \pm 0.3$     \\
    {\textsc{DualSort}, Adap.}        & ${ 10.3 \pm 0.1}$   & $ 5.27 \pm 0.3$     \\
    {\textsc{DualSort}, Ande.}        & $\mathbf{8.2 \pm 0.1}$ & $\mathbf{ 3.72  \pm 0.3}$   \\
    \bottomrule
    \end{tabular}
    }}
    \end{table}
    We follow the experimental setup from \citep{cuturi2019differentiable}. The classifier network from \citep{cuturi2022optimal} is used for CIFAR-100, consisting of four CNN layers, and a fully connected hidden layer, full details given in \S\ref{sec:app_soft_err}.
    
    The $\varepsilon$ regularization was set to $0.01$ and the network was trained until convergence over $10$ seeds. \textsc{DualSort} initializer was ran with $3$ iterations, which, as discussed in \S\ref{subsec:sorting}, is slightly cheaper than two Sinkhorn iterations.
    
    Accuracy on the evaluation set is shown in \Cref{fig:cifar_classifier} for $300$ epochs. It is clear that, as expected, the Sinkhorn initialization procedure does not affect training nor accuracy. However, \Cref{tab:soft_error} shows that the \textsc{DualSort} initializer drastically reduces the number of Sinkhorn iterations needed for convergence, to compute the soft-error loss and its gradients at each evaluation.

\subsection{Differentiable Clustering}\label{sec:clustering}
    We demonstrate the performance improvement from the Gaussian initializer on the task of deep differentiable clustering, with the experimental setup of \citep{genevay2018learning}. Differentiable clustering aims at producing a latent representation amenable to clustering. This is achieved using a variational autoencoder \citep{kingma2014semi} with learnable, discrete cluster embeddings, and an additional loss term allocating encodings to cluster embeddings using OT. 
\begin{table}[h]
\centering
{\small{\centering
\caption{Avg. Sinkhorn iter./training step and runtime / training step $\textrm{mean} \pm \textrm{std}$ for differentiable clustering VAE, $10$ seeds, $\epsilon=0.001$}
\label{tab:diff_clust_iter}
\begin{tabular}{lll}
    \toprule
    {}                   & Iterations & Runtime ($\times 10^{-3}$ s)  \\ \midrule
    {Zero}            & $354.1 \pm 7.0$    & ${ 25.4 \pm 0.2 }$  \\
    {$\varepsilon$-decay}       &  ${340.5 \pm 17.8}$  & ${ 25.1 \pm 0.1 }$    \\
    {Anderson}        &  ${844.4 \pm 26.2}$  & ${ 144 \pm 6.7 }$    \\
    {Momentum}        &  ${342.5 \pm 3.7}$  & ${ 33.1  \pm 1.7 }$    \\
    {Adaptive}        &  ${96.6 \pm 4.1}$   & ${ 9.35 \pm 0.02 }$  \\
    {Gaus}            &  ${196.6\pm 6.7}$   & ${ 16.2\pm 0.6  }$   \\
    {Gaus, Adapt.}    &  $\mathbf{68.7\pm 1.3}$  & $\mathbf{ 8.00 \pm 0.1 }$    \\
        \bottomrule
    \end{tabular}}}
    \end{table}

    For data of dimension $d_x$ and latent dimension $d_z$, let $E_\theta: \mathbb{R}^{d_x} \rightarrow \mathbb{R}^{2 \times d_z}$ and $D_\theta:\mathbb{R}^{d_z} \rightarrow \mathbb{R}^{d_x}$ denote an encoder and decoder respectively, parameterized by $\theta$. Let $\mu_\phi \in \mathbb{R}^{K \times d_z}$ denote cluster embeddings for $K$ clusters. The objective of differentiable clustering is to learn $E_\theta, D_\theta$ and embeddings $\mu_\phi \in \mathbb{R}^{K \times d_z}$. This may be achieved by minimizing the loss $\ell^\textrm{ae}(\theta) + \ell^\textrm{OT}(\phi, \theta)$ for each batch of data $(x_i)_i$. Here $\ell^\textrm{ae}(\theta)$ is the standard variational auto-encoder loss and $\ell^\textrm{OT}(\phi, \theta)$ is the regularized OT loss from \eqref{eq:erot} between $\mu=\sum_{k=1}^{K}\tfrac{1}{K}\delta_{\mu_{\phi k}}$ and $\nu=\sum_{i=1}^{n}\tfrac{1}{n}\delta_{z_i}$. $z_i = \mathbf{m}_i + \sigma_i u_i$, where $(\mathbf{m}_i, \sigma_i) = E_\theta(x_i)$, $u_i\sim \mathcal{N}(\mathbf{0}_{d_z}, \mathbf{I}_{d_z})$,  and $\tilde{x}_i=D_\theta(z_i)$.

    We demonstrate this task for MNIST \citep{deng2012mnist} over $10$ seeds. Fully connected networks with $4$ hidden layers were used for $E_\theta$ and $D_\theta$, where $d_z=32$ and $d_x=784$, further experimental details are given in \S\ref{sec:app_clus}. \Cref{tab:diff_clust_iter} shows that the Gaussian initializer outperforms the zero initialization for default Sinkhorn and all other combinations of default Sinkhorn plus acceleration techniques. 
    Performance metrics and samples from the generative model are given in Appendix \ref{sec:app_clus}.

\subsection{Differentiable Particle Filtering} \label{sec:dpf}
As introduced in \citet{corenflos2021differentiable}, the Sinkhorn algorithm provides an approximate differentiable resampling scheme, hence enables end-to-end differentiable particle filtering. Consider a simple linear state space model consisting of latent states $x_t\in \mathbb{R}^2$ where $x_0 = \mathbf{0}$, $X_t|x_{t-1} \sim f(\cdot| x_{t-1}) = \mathcal{N}(0.5\mathbb{I} x_{t-1}, \mathbb{I})$ and observations $y_t\in \mathbb{R}^2$, $y_t\sim g(\cdot| x_{t}) =\mathcal{N}(x_{t}, \mathbb{I})$ for $t \in \{1, \ldots, T\}$, and time series length $T=500$.  Differentiable resampling via OT consists of applying the Sinkhorn algorithm between weighted and unweighted pointclouds of $N$ simulated latent states at each timepoint $t$, for each forward pass. For full details see \citet{corenflos2021differentiable}.

\begin{table}[ht]
\caption{Mean $\pm$ std number of Sinkhorn iterations and runtime over $3$ seeds for the forward pass of a particle filter with $N$ particles, batch size $4$ of simple linear state space model, $T=500$.}
\label{tab:pf_res}
\centering
{\small
\begin{tabular}{llll}
 \textbf{N}       & \textbf{Initializer}     & \textbf{Iterations ('000s)} & \textbf{Runtime /s} \\ \hline
$32$  & Gaus                & $440  \pm 2.5$                      & $12.08 \pm 0.25$    \\
      & $\mathbf{0}$                    & $611 \pm 3.4$                       & $15.46 \pm 0.35$    \\ \hline
$64$  & Gaus                & $349 \pm 2.9$                       & $9.62 \pm 1.29$     \\
      & $\mathbf{0}$                    & $532 \pm 3.4$                       & $12.49 \pm 0.69$    \\ \hline
$128$ & Gaus                & $269 \pm 0.7$                       & $7.03 \pm 1.21$     \\
      & $\mathbf{0}$                & $471 \pm 2.3$                       & $10.18 \pm 0.88$    \\ \hline
$256$ & Gaus                & $216 \pm 1.5$                       & $6.34 0.78$         \\
      & $\mathbf{0}$                    & $439 \pm 1.9$                       & $11.01 \pm 0.59$    \\ \hline
$512$ & Gaus                & $176 \pm 1.3$             & $14.43 \pm 1.40$    \\
      & $\mathbf{0}$ & $422 \pm 1.7$                       & $30.17 \pm 1.06$    \\
      \hline
\end{tabular}
}
\end{table}

 For batch size $B=4$ and time steps $T =500$, under a naive implementation, each forward pass requires $T \times B$  Sinkhorn layer evaluations. This can be slow. As shown in \Cref{tab:pf_res}, the Gaussian initializer is effective at reducing the runtime by reducing the number of Sinkhorn iterations by approximately $33\%$ to $50\%$ relative to the default Sinkhorn with $\mathbf{0}$ initialization.
 
\subsection{Document Similarity}\label{sec:word_embeddings}
In this experiment, we compare the \textsc{Gaus}, \textsc{GMM} and \textsc{Subsample} initializers. Documents were gathered from the $20$ \textit{Newsgroup} dataset \citep{newsgroup} and each word, $(w_i)_{i=1}^n$, in the vocabulary across documents is embedded using the pre-trained GloVe word embeddings \citep{pennington2014glove} as $(e_i)_{i=1}^n$ where $e_i \in \mathbb{R}^{50}$.

    \begin{figure}[ht]
    \centering
    \includegraphics[trim= 0 30 0 80 , clip , width=\linewidth]{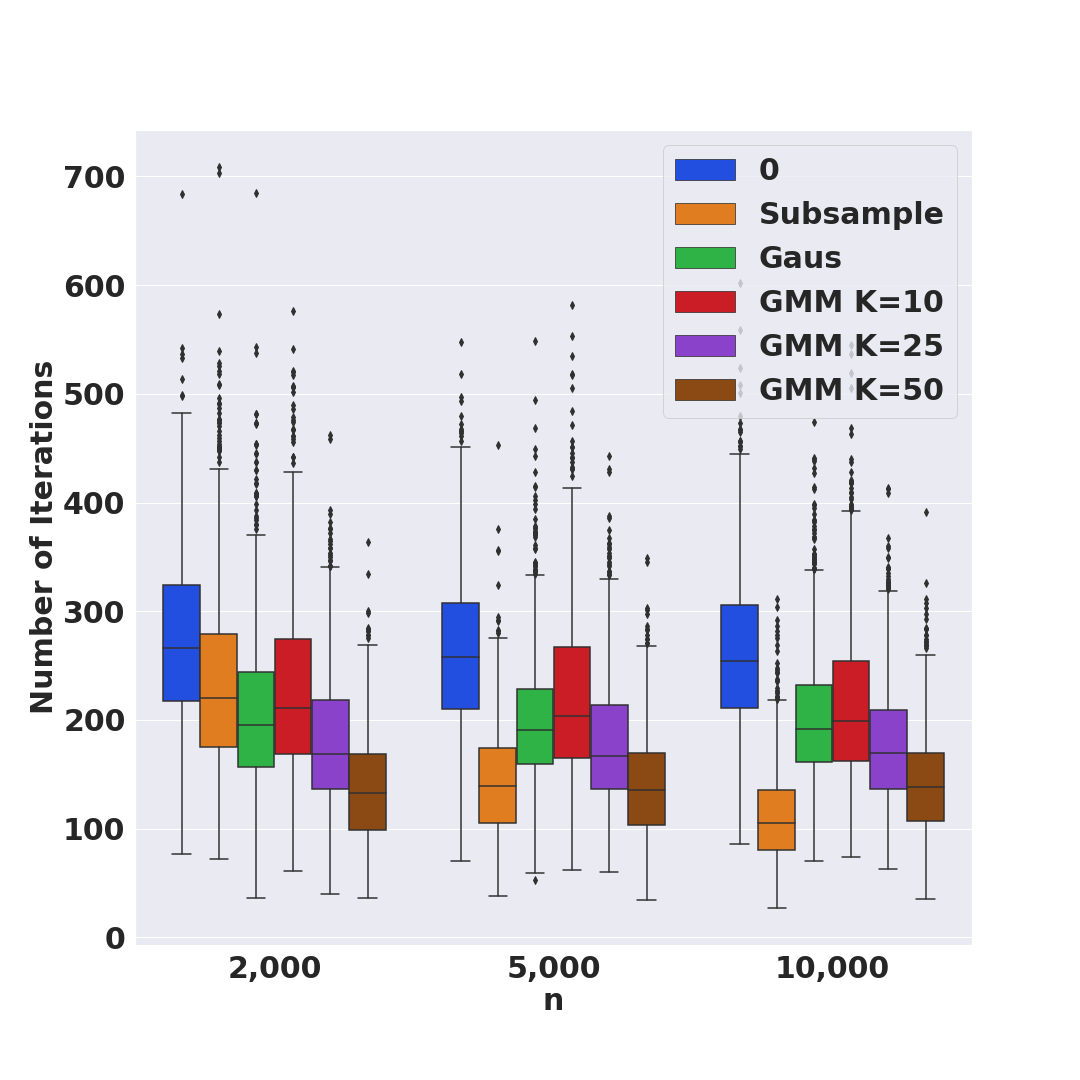}
    \caption{Distribution of number of Sinkhorn iterations required for Sinkhorn convergence between 1225 pairs of Newsgroup documents, represented as word embeddings histograms, $n$ being the total vocabulary size. The same convergence threshold for Sinkhorn is used for all $n$.}
    \label{fig:word_emb}
    \end{figure} 
     In a similar setup to \citet{kusner2015word}, each document may be represented as a histogram with weights $(a_i)_{i=1}^n$ corresponding to word-frequency, $\nu_i = \sum_{i=1}^n a_i\delta_{e_i}$, and we compute pairwise OT distances between $50$ documents, resulting in $1,225$ pairs. We report the number of Sinkhorn iterations and runtime required for convergence for the default zero intializer ($\mathbf{0}$), the proposed \textsc{Gaus} initializer, the \textsc{GMM} initializers with full covariance matrices and $K \in \{10, 25, 50\}$ components, and the \textsc{Subsample} initializer. A subset of the vocabulary of size $n\in \{2\times10^3, 5\times10^3, 10^4\}$ was used, and corresponding subsample of size $100, 500$ and $1,000$ for the \textsc{Subsample} initializer. Regularization was $\varepsilon=0.001$.
     
\begin{figure}[h]
    \centering
    \includegraphics[width=\linewidth]{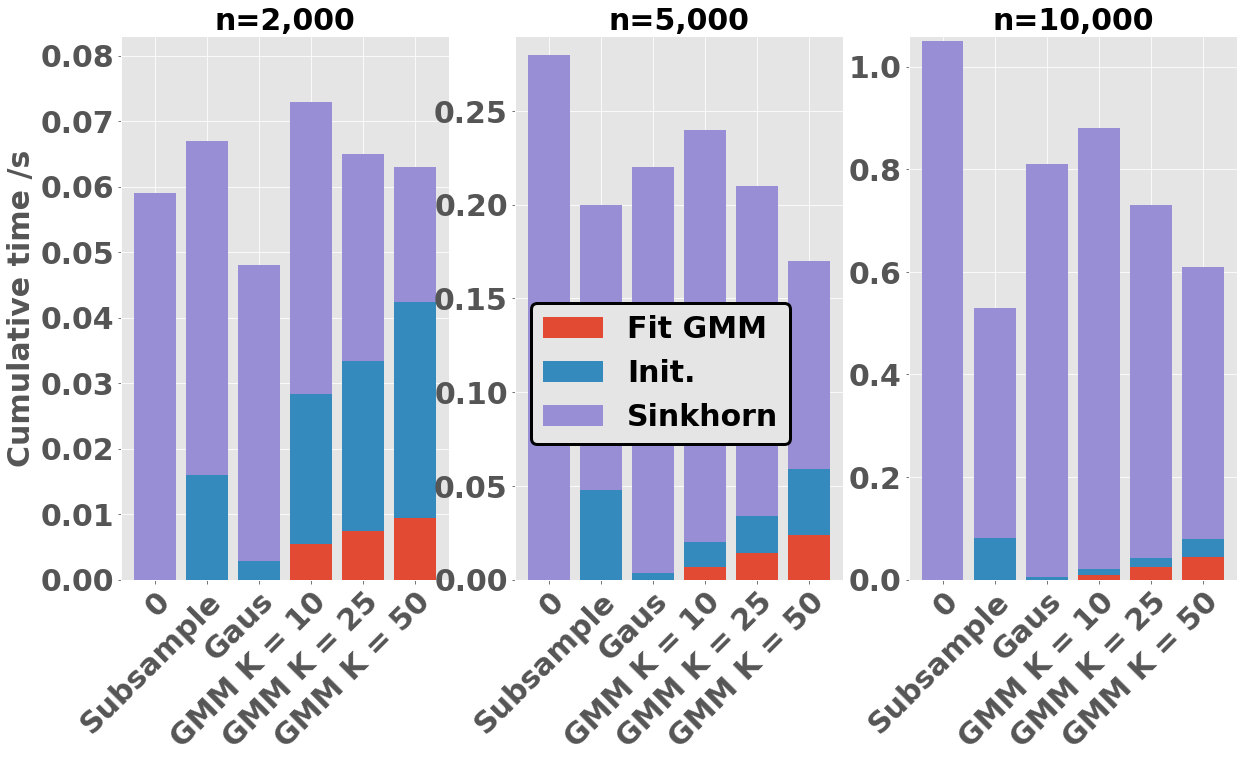}
    \caption{Average wall clock time for computing OT between each pair of word-embeddings ($1,225$ problems) for vocabulary of size $n=2 \times 10^3; 5\times 10^3; 10^4$; split by initialization time (Init), time to compute Gaussian mixture models (Fit GMM) and Sinkhorn iterations (Sinkhorn).}
    \label{fig:word_emb_timing}
    \end{figure}

     The distribution of results are shown in \Cref{fig:word_emb} and \Cref{fig:word_emb_timing} illustrating that improvements can be obtained for a range of $K$. Notice however that \textsc{Gaus} beats the \textsc{GMM} for low $K$, we suspect this is due to the additional approximation \eqref{eq:gmm_approx}. Although often resulting in lower number of fine-tuning Sinkhorn iterations, the preprocessing cost of running the \textsc{Subsample} initializer is expensive, and only exhibits better aggregate runtime performance for large $n=10,000$, which was expected. A GMM was first fitted to each document, before being used for initializing Sinkhorn potentials. As \Cref{fig:word_emb_timing} shows, although the cost of fitting GMMs results in limited runtime savings for $n=2,000$, there are significant runtime savings for $n=5,000$ and $n=10,000$. See further discussion in \S\ref{sec:overhead}.

\section{Conclusion} 
We have introduced efficient and robust Sinkhorn potential initialization schemes: \textsc{DualSort}, \textsc{Gaus}, \textsc{GMM} and demonstrated how these carefully chosen initializers can significantly improve the performance of the Sinkhorn algorithm for a variety of tasks. These GPU-friendly initializers may also be embedded in end-to-end differentiable procedures by relying on implicit differentiation, as demonstrated in various tasks presented in our experiments (ranking, clustering, filtering), and are complementary to most common acceleration methods, creating an interesting space to optimize further the execution of Sinkhorn. Initialization is a neglected area of computational OT, and we hope that these promising results can inspire new research to other areas, such as initalizing calls to Sinkhorn in the internal loops of the Gromov-Wasserstein or barycenter problem. We also hope they can help extending OT's reach to data-hungry application areas, such as single-cell or NLP tasks that involve typically a large number of samples.

\bibliographystyle{apalike}
\bibliography{references}

\begin{thebibliography}{}

\bibitem[Adams and Zemel, 2011]{adams2011ranking}
Adams, R.~P. and Zemel, R.~S. (2011).
\newblock Ranking via sinkhorn propagation.
\newblock {\em arXiv preprint arXiv:1106.1925}.

\bibitem[Ahuja et~al., 1988]{ahuja1988network}
Ahuja, R.~K., Magnanti, T.~L., and Orlin, J.~B. (1988).
\newblock Network flows.

\bibitem[Altschuler et~al., 2019]{altschuler2019massively}
Altschuler, J., Bach, F., Rudi, A., and Niles-Weed, J. (2019).
\newblock Massively scalable sinkhorn distances via the nystr{\"o}m method.
\newblock {\em Advances in neural information processing systems}, 32.

\bibitem[Amos et~al., 2022]{amos22}
Amos, B., Cohen, S., Luise, G., and Redko, I. (2022).
\newblock Meta optimal transport.

\bibitem[Anderson, 1965]{anderson1965iterative}
Anderson, D.~G. (1965).
\newblock Iterative procedures for nonlinear integral equations.
\newblock {\em Journal of the ACM (JACM)}, 12(4):547--560.

\bibitem[Bertsimas and Tsitsiklis, 1997]{bertsimas1997introduction}
Bertsimas, D. and Tsitsiklis, J.~N. (1997).
\newblock {\em Introduction to linear optimization}, volume~6.
\newblock Athena Scientific Belmont, MA.

\bibitem[Bradbury et~al., 2018]{bradbury2018jax}
Bradbury, J., Frostig, R., Hawkins, P., Johnson, M.~J., Leary, C., Maclaurin,
  D., Necula, G., Paszke, A., VanderPlas, J., Wanderman-Milne, S., et~al.
  (2018).
\newblock Jax: composable transformations of python+ numpy programs.
\newblock {\em Version 0.2}, 5:14--24.

\bibitem[Brenier, 1987]{brenier1987decomposition}
Brenier, Y. (1987).
\newblock D{\'e}composition polaire et r{\'e}arrangement monotone des champs de
  vecteurs.
\newblock {\em CR Acad. Sci. Paris S{\'e}r. I Math.}, 305:805--808.

\bibitem[Caron et~al., 2020]{caron2020unsupervised}
Caron, M., Misra, I., Mairal, J., Goyal, P., Bojanowski, P., and Joulin, A.
  (2020).
\newblock Unsupervised learning of visual features by contrasting cluster
  assignments.
\newblock {\em Advances in Neural Information Processing Systems},
  33:9912--9924.

\bibitem[Chen et~al., 2018]{chen2018optimal}
Chen, Y., Georgiou, T.~T., and Tannenbaum, A. (2018).
\newblock Optimal transport for gaussian mixture models.
\newblock {\em IEEE Access}, 7:6269--6278.

\bibitem[Chiappori et~al., 2017]{chiappori2017multi}
Chiappori, P.-A., McCann, R.~J., and Pass, B. (2017).
\newblock Multi-to one-dimensional optimal transport.
\newblock {\em Communications on Pure and Applied Mathematics},
  70(12):2405--2444.

\bibitem[Chizat et~al., 2020]{chizat2020faster}
Chizat, L., Roussillon, P., L{\'e}ger, F., Vialard, F.-X., and Peyr{\'e}, G.
  (2020).
\newblock Faster wasserstein distance estimation with the sinkhorn divergence.
\newblock {\em Advances in Neural Information Processing Systems},
  33:2257--2269.

\bibitem[Corenflos et~al., 2021]{corenflos2021differentiable}
Corenflos, A., Thornton, J., Deligiannidis, G., and Doucet, A. (2021).
\newblock Differentiable particle filtering via entropy-regularized optimal
  transport.
\newblock In {\em International Conference on Machine Learning}, pages
  2100--2111. PMLR.

\bibitem[Courty et~al., 2017]{courty2017joint}
Courty, N., Flamary, R., Habrard, A., and Rakotomamonjy, A. (2017).
\newblock Joint distribution optimal transportation for domain adaptation.
\newblock {\em Advances in Neural Information Processing Systems}, 30.

\bibitem[Courty et~al., 2014]{courty2014domain}
Courty, N., Flamary, R., and Tuia, D. (2014).
\newblock Domain adaptation with regularized optimal transport.
\newblock In {\em Joint European Conference on Machine Learning and Knowledge
  Discovery in Databases}, pages 274--289. Springer.

\bibitem[Cuturi, 2013]{cuturi2013sinkhorn}
Cuturi, M. (2013).
\newblock Sinkhorn distances: Lightspeed computation of optimal transport.
\newblock {\em Advances in neural information processing systems}, 26.

\bibitem[Cuturi et~al., 2022]{cuturi2022optimal}
Cuturi, M., Meng-Papaxanthos, L., Tian, Y., Bunne, C., Davis, G., and Teboul,
  O. (2022).
\newblock Optimal transport tools (ott): A jax toolbox for all things
  wasserstein.
\newblock {\em arXiv preprint arXiv:2201.12324}.

\bibitem[Cuturi and Peyr{\'e}, 2015]{cuturi2015smoothed}
Cuturi, M. and Peyr{\'e}, G. (2015).
\newblock A smoothed dual approach for variational wasserstein problems.
\newblock {\em arXiv preprint arXiv:1503.02533}.

\bibitem[Cuturi et~al., 2020]{cuturi2020supervised}
Cuturi, M., Teboul, O., Niles-Weed, J., and Vert, J.-P. (2020).
\newblock Supervised quantile normalization for low rank matrix factorization.
\newblock In {\em International Conference on Machine Learning}, pages
  2269--2279. PMLR.

\bibitem[Cuturi et~al., 2019]{cuturi2019differentiable}
Cuturi, M., Teboul, O., and Vert, J.-P. (2019).
\newblock Differentiable ranking and sorting using optimal transport.
\newblock {\em Advances in neural information processing systems}, 32.

\bibitem[Dantzig et~al., 1956]{dantzig1956primal}
Dantzig, G.~B., Ford~Jr, L.~R., and Fulkerson, D.~R. (1956).
\newblock A primal--dual algorithm.
\newblock Technical report, RAND CORP SANTA MONICA CA.

\bibitem[Deng, 2012]{deng2012mnist}
Deng, L. (2012).
\newblock The mnist database of handwritten digit images for machine learning
  research.
\newblock {\em IEEE Signal Processing Magazine}, 29(6):141--142.

\bibitem[d’Aspremont et~al., 2021]{d2021acceleration}
d’Aspremont, A., Scieur, D., Taylor, A., et~al. (2021).
\newblock Acceleration methods.
\newblock {\em Foundations and Trends{\textregistered} in Optimization},
  5(1-2):1--245.

\bibitem[Feydy, 2020]{feydy2020geometric}
Feydy, J. (2020).
\newblock {\em Geometric data analysis, beyond convolutions}.
\newblock PhD thesis, Universit{\'e} Paris-Saclay Gif-sur-Yvette, France.

\bibitem[Flamary et~al., 2021]{flamary2021pot}
Flamary, R., Courty, N., Gramfort, A., Alaya, M.~Z., Boisbunon, A., Chambon,
  S., Chapel, L., Corenflos, A., Fatras, K., Fournier, N., et~al. (2021).
\newblock Pot: Python optimal transport.
\newblock {\em Journal of Machine Learning Research}, 22(78):1--8.

\bibitem[Genevay et~al., 2019]{genevay2019differentiable}
Genevay, A., Dulac-Arnold, G., and Vert, J.-P. (2019).
\newblock Differentiable deep clustering with cluster size constraints.
\newblock {\em arXiv preprint arXiv:1910.09036}.

\bibitem[Genevay et~al., 2018]{genevay2018learning}
Genevay, A., Peyr{\'e}, G., and Cuturi, M. (2018).
\newblock Learning generative models with sinkhorn divergences.
\newblock In {\em International Conference on Artificial Intelligence and
  Statistics}, pages 1608--1617. PMLR.

\bibitem[Hashimoto et~al., 2016]{hashimoto2016learning}
Hashimoto, T., Gifford, D., and Jaakkola, T. (2016).
\newblock {Learning Population-Level Diffusions with Generative Recurrent
  Networks}.
\newblock volume~33.

\bibitem[Higham, 2008]{higham2008functions}
Higham, N.~J. (2008).
\newblock {\em Functions of matrices: theory and computation}.
\newblock SIAM.

\bibitem[Janati et~al., 2020]{janati2020multi}
Janati, H., Bazeille, T., Thirion, B., Cuturi, M., and Gramfort, A. (2020).
\newblock Multi-subject meg/eeg source imaging with sparse multi-task
  regression.
\newblock {\em NeuroImage}, 220:116847.

\bibitem[Kingma et~al., 2014]{kingma2014semi}
Kingma, D.~P., Mohamed, S., Jimenez~Rezende, D., and Welling, M. (2014).
\newblock Semi-supervised learning with deep generative models.
\newblock {\em Advances in neural information processing systems}, 27.

\bibitem[Kosowsky and Yuille, 1994]{kosowsky1994invisible}
Kosowsky, J. and Yuille, A.~L. (1994).
\newblock The invisible hand algorithm: Solving the assignment problem with
  statistical physics.
\newblock {\em Neural networks}, 7(3):477--490.

\bibitem[Kusner et~al., 2015]{kusner2015word}
Kusner, M., Sun, Y., Kolkin, N., and Weinberger, K. (2015).
\newblock From word embeddings to document distances.
\newblock In {\em International conference on machine learning}, pages
  957--966. PMLR.

\bibitem[Lang, 1995]{newsgroup}
Lang, K. (1995).
\newblock Newsweeder: Learning to filter netnews.
\newblock In {\em Proceedings of the Twelfth International Conference on
  Machine Learning}, pages 331--339.

\bibitem[Lehmann et~al., 2021]{lehmann2021note}
Lehmann, T., Von~Renesse, M.-K., Sambale, A., and Uschmajew, A. (2021).
\newblock A note on overrelaxation in the sinkhorn algorithm.
\newblock {\em Optimization Letters}, pages 1--12.

\bibitem[Luise et~al., 2018]{luise2018differential}
Luise, G., Rudi, A., Pontil, M., and Ciliberto, C. (2018).
\newblock Differential properties of sinkhorn approximation for learning with
  wasserstein distance.
\newblock {\em Advances in Neural Information Processing Systems}, 31.

\bibitem[Meyron, 2019]{MEYRON201913}
Meyron, J. (2019).
\newblock Initialization procedures for discrete and semi-discrete optimal
  transport.
\newblock {\em Computer-Aided Design}, 115:13--22.

\bibitem[Monge, 1781]{Monge1781}
Monge, G. (1781).
\newblock M{\'e}moire sur la th{\'e}orie des d{\'e}blais et des remblais.
\newblock {\em Histoire de l'Acad{\'e}mie Royale des Sciences}, pages 666--704.

\bibitem[Pedregosa et~al., 2011]{scikit_learn}
Pedregosa, F., Varoquaux, G., Gramfort, A., Michel, V., Thirion, B., Grisel,
  O., Blondel, M., Prettenhofer, P., Weiss, R., Dubourg, V., Vanderplas, J.,
  Passos, A., Cournapeau, D., Brucher, M., Perrot, M., and Duchesnay, E.
  (2011).
\newblock Scikit-learn: Machine learning in {P}ython.
\newblock {\em Journal of Machine Learning Research}, 12:2825--2830.

\bibitem[Pennington et~al., 2014]{pennington2014glove}
Pennington, J., Socher, R., and Manning, C.~D. (2014).
\newblock Glove: Global vectors for word representation.
\newblock In {\em Empirical Methods in Natural Language Processing (EMNLP)},
  pages 1532--1543.

\bibitem[Peyr{\'e} et~al., 2019]{peyre2019computational}
Peyr{\'e}, G., Cuturi, M., et~al. (2019).
\newblock Computational optimal transport: With applications to data science.
\newblock {\em Foundations and Trends{\textregistered} in Machine Learning},
  11(5-6):355--607.

\bibitem[Pooladian and Niles-Weed, 2021]{pooladian}
Pooladian, A.-A. and Niles-Weed, J. (2021).
\newblock Entropic estimation of optimal transport maps.

\bibitem[Salimans et~al., 2018]{salimans2018improving}
Salimans, T., Zhang, H., Radford, A., and Metaxas, D. (2018).
\newblock Improving {GAN}s using optimal transport.
\newblock In {\em International Conference on Learning Representations}.

\bibitem[Sander et~al., 2022]{sander2022sinkformers}
Sander, M.~E., Ablin, P., Blondel, M., and Peyr{\'e}, G. (2022).
\newblock Sinkformers: Transformers with doubly stochastic attention.
\newblock In {\em International Conference on Artificial Intelligence and
  Statistics}, pages 3515--3530. PMLR.

\bibitem[Santambrogio, 2015]{SantambrogioBook}
Santambrogio, F. (2015).
\newblock {\em Optimal transport for applied mathematicians}.
\newblock Birkhauser.

\bibitem[Sarlin et~al., 2020]{Sarlin_2020_CVPR}
Sarlin, P.-E., DeTone, D., Malisiewicz, T., and Rabinovich, A. (2020).
\newblock Superglue: Learning feature matching with graph neural networks.
\newblock In {\em Proceedings of the IEEE/CVF Conference on Computer Vision and
  Pattern Recognition (CVPR)}.

\bibitem[Scetbon and Cuturi, 2020]{scetbon2020linear}
Scetbon, M. and Cuturi, M. (2020).
\newblock Linear time sinkhorn divergences using positive features.
\newblock {\em Advances in Neural Information Processing Systems},
  33:13468--13480.

\bibitem[Schiebinger et~al., 2019]{schiebinger2019optimal}
Schiebinger, G., Shu, J., Tabaka, M., Cleary, B., Subramanian, V., Solomon, A.,
  Gould, J., Liu, S., Lin, S., Berube, P., et~al. (2019).
\newblock {Optimal-Transport Analysis of Single-Cell Gene Expression Identifies
  Developmental Trajectories in Reprogramming}.
\newblock {\em Cell}, 176(4).

\bibitem[Schmitz et~al., 2018]{schmitz2018wasserstein}
Schmitz, M.~A., Heitz, M., Bonneel, N., Ngole, F., Coeurjolly, D., Cuturi, M.,
  Peyr{\'e}, G., and Starck, J.-L. (2018).
\newblock Wasserstein dictionary learning: Optimal transport-based unsupervised
  nonlinear dictionary learning.
\newblock {\em SIAM Journal on Imaging Sciences}, 11(1):643--678.

\bibitem[Schmitzer, 2019]{schmitzer2019stabilized}
Schmitzer, B. (2019).
\newblock Stabilized sparse scaling algorithms for entropy regularized
  transport problems.
\newblock {\em SIAM Journal on Scientific Computing}, 41(3):A1443--A1481.

\bibitem[Sejourne et~al., 2022]{pmlr-v151-sejourne22a}
Sejourne, T., Vialard, F.-X., and Peyr\'e, G. (2022).
\newblock Faster unbalanced optimal transport: Translation invariant sinkhorn
  and 1-d frank-wolfe.
\newblock In Camps-Valls, G., Ruiz, F. J.~R., and Valera, I., editors, {\em
  Proceedings of The 25th International Conference on Artificial Intelligence
  and Statistics}, volume 151 of {\em Proceedings of Machine Learning
  Research}, pages 4995--5021. PMLR.

\bibitem[Sinkhorn, 1967]{Sinkhorn67}
Sinkhorn, R. (1967).
\newblock Diagonal equivalence to matrices with prescribed row and column sums.
\newblock {\em American Mathematical Monthly}, 74:402--405.

\bibitem[Solomon et~al., 2015]{2015-solomon-siggraph}
Solomon, J., De~Goes, F., Peyr{\'e}, G., Cuturi, M., Butscher, A., Nguyen, A.,
  Du, T., and Guibas, L. (2015).
\newblock Convolutional {Wasserstein} distances: efficient optimal
  transportation on geometric domains.
\newblock {\em ACM Transactions on Graphics}, 34(4):66:1--66:11.

\bibitem[Thibault et~al., 2021]{thibault2021overrelaxed}
Thibault, A., Chizat, L., Dossal, C., and Papadakis, N. (2021).
\newblock Overrelaxed sinkhorn--knopp algorithm for regularized optimal
  transport.
\newblock {\em Algorithms}, 14(5):143.

\bibitem[Xie et~al., 2020a]{NEURIPS2020_ec24a54d}
Xie, Y., Dai, H., Chen, M., Dai, B., Zhao, T., Zha, H., Wei, W., and Pfister,
  T. (2020a).
\newblock Differentiable top-k with optimal transport.
\newblock In Larochelle, H., Ranzato, M., Hadsell, R., Balcan, M., and Lin, H.,
  editors, {\em Advances in Neural Information Processing Systems}, volume~33,
  pages 20520--20531. Curran Associates, Inc.

\bibitem[Xie et~al., 2020b]{xie2020fast}
Xie, Y., Wang, X., Wang, R., and Zha, H. (2020b).
\newblock A fast proximal point method for computing exact wasserstein
  distance.
\newblock In {\em Uncertainty in artificial intelligence}, pages 433--453.
  PMLR.

\end{thebibliography}

\newpage
\appendix

{
\onecolumn

\section{Dual Potential Comparison}\label{sec:app_dual}
For balanced OT problems, as considered here, Dual potentials $\bff$, $\bg$ are unique up to constant shifts i.e. $\bff-s$, $\bg+s$ for $s \in \mathbb{R}$. Therefore, in order to compare potentials $\bff \in \mathbb{R}^n$, we center $\bff$, as $\bff \gets \bff - \frac{1}{n}\sum_i \bff_i$.

\subsection{From Optimal Primal to Dual Vectors}\label{sec:primaldual}
\textbf{Properties of the optimal primal $\bP^\star$. } Taking the 1D case as motivation, we introduce a method to recover optimal dual potentials $\mathbf{f^\star,g^\star}$ from an optimal primal solution $\bP^\star$. To that end, one can cast an OT problem as a min-cost-flow problem on a bipartite graph $\mathcal{G}=(\mathcal{V}, \mathcal{E})$, with vertices composed of source nodes $S=\{1,\ldots,n\}$ and target nodes $T=\{1', \ldots, m'\}$, $\mathcal{V}=S\cup T$, and edge set $\mathcal{E}=\{(i,j'), i=1,\dots,n; j=1,\dots,m\}$ linking them. The KKT conditions state that, writing $\mathcal{E}(\bP)=\{(i,j') | \bP_{i,j}>0\}$ one has that the graph $(\mathcal{V},\mathcal{E}(\bP^\star))$ is necessarily a forest~\citep[Prop. 3.4]{peyre2019computational}.

We write $\mathcal{T}_1, \dots, \mathcal{T}_K$ for the $K$ trees forming that forest, where $1\leq K \leq \min(n, m)$, and write $t_k$ for their size. 

We use the lexicographic order to define the root node of each tree, chosen to be the smallest \textit{source} node $s(k)$ contained in $\mathcal{T}_k$. For convenience, we assume that trees are ordered following $s(k)$, and therefore that $\mathcal{T}_1$ has $1$ as its root node.
For each tree $k$, we introduce $p^k=(p^k_1,\dots, p^k_{t_k-1})$ for a pre-order breadth-first-traversal of $\mathcal{T}_k$ originating at $s(k)$, enumerating $t_k-1$ edges, namely pairs in $S\times T$ or $T\times S$, guaranteed to be such that any parent node in the tree is visited before its descendants.
$\iota(j)$ denotes the smallest source index $i$ such that $(i,j')\in\mathcal{E}(\bP^\star)$.

\begin{algorithm}[h]
        \caption{Recover dual from primal}
        \label{algo:primal_init}
        \begin{algorithmic}[1]
            \STATE{{\bfseries Input:} Cost matrix $\mathbf{C}$ and graph $(\mathcal{V},\mathcal{E}(\bP^\star))$}
            \STATE{\bfseries Initialize:} $\bff = 0$.
            \WHILE{not converged}
                  \FOR{$k\in \{2,\dots, K\}$}
                    \STATE $\bff_{s(k)} \leftarrow \min_j c_{s(k),j}-c_{\iota(j), j}+\bff_{\iota(j)}$
                  \ENDFOR
                  \FOR{$k\in \{1,\dots, K\}$}    
                    \STATE $\bff \gets \textsc{UpdateTree}(\mathbf{C}, \bff, k)$
                  \ENDFOR
            \ENDWHILE
            \STATE{{\bfseries Return} {$\mathbf{f}$}}
        \end{algorithmic}
        \end{algorithm}
        \begin{algorithm}[h]
        \caption{\textsc{UpdateTree}}
        \label{algo:update_tree}
        \begin{algorithmic}[1]
            \STATE{{\bfseries Input:} Cost matrix $\mathbf{C}$, $\bff$, tree index $k$}
            \FOR{$e=(a,b) \in p^k$}
             \IF{$a\in S, b\in T$}
                \STATE $g \leftarrow c_{a,b}-\bff_{a}$
             \ELSE
                \STATE $\bff_a \leftarrow c_{a,b}-g$
             \ENDIF
            \ENDFOR
            \STATE{{\bfseries Return} {$\mathbf{f}$}}
        \end{algorithmic}
        \end{algorithm}

\textbf{Complementary and Feasibility Constraints. } Complementary slackness provides a set of ${n+m-K}$ linear equations \eqref{eq:comp_condition}, while feasibility constraints are given in \eqref{eq:feas_condition}.
\begin{equation}
(i,j')\in\mathcal{E}(\bP^\star) \Leftrightarrow \bff_i^\star+\bg_j^\star = c_{i,j}\;, 
\label{eq:comp_condition}
\end{equation}
\begin{equation}
\forall i\leq n,j \leq m,\,\, \bff_i+\bg_j \leq c_{i,j}\,. \label{eq:feas_condition}
\end{equation}
For the special case $K=1$, which happens for instance when $n$ and $m$ are co-primes and weights are uniform, the set of linear equations  \eqref{eq:comp_condition} suffices to recover the $n+m$ dual variables, with the convention that the first entry be $0$. When, on the contrary, $K>1$, that set of $n+m-K$ equations is no longer sufficient. For example, $K=n=m$ for the optimal assignment problem, in which $(\mathcal{V},\mathcal{E}(\bP^\star))$ describes a set of $n$ isolated trees, and only $n$ equality relations are available for $2n$ variables.  In such cases, one must additionally use the feasibility constraint \eqref{eq:feas_condition} to obtain optimal dual variables \citep[Prop 3.3]{peyre2019computational}.

 The $c$-transform ${\bg^c_i := \min_j c_{i,j}-\bg_j}$ can be used to enforce constraints \eqref{eq:feas_condition}, however, it may no longer satisfy the complementary condition \eqref{eq:comp_condition}. This is remedied by updating all source nodes $i$ in tree $k$ by starting from $s(k)$ as detailed in \Cref{algo:update_tree}. Repeated application of these updates, \Cref{algo:primal_init}, guarantees convergence.

\begin{lemma}\label{lem:conv_oned}
     Given the optimal coupling matrix $\mathbf{P}^*$ solving OT problem \eqref{eq:erot} with $\varepsilon=0$, the procedure defined in \Cref{algo:primal_init} converges to the optimal dual potentials for dual problem \eqref{eq:ot_dual}.
\end{lemma}

The proof is provided in \S\ref{sec:dualsort_conv}, and uses the fact that \Cref{algo:primal_init} is a primal-dual method \citep{dantzig1956primal}, tweaked because the primal solution $\bP^\star$ is known.

\section{Further Experimental Details}\label{sec:app_exp}

\subsection{Differentiable Sorting Details}\label{sec:app_softsort}
Regularization $\varepsilon=0.01$ was used, as per \citep{cuturi2019differentiable}. In this experiment arrays of size $n \in \{16, 32, 64, 128, 256, 512, 1024\}$ were sampled from the Gaussian blob dataset \citep{scikit_learn} for $200$ different seeds. At each seed, $1$-dimensional Gaussian data is generated from $5$ random centers with centers uniformly distributed in $(-10, 10)$ with standard deviation $3$.

Baseline acceleration methods (Anderson acceleration, momentum, adaptive momentum, $\epsilon$ decay) were considered to augment the Sinkhorn algorithm, using the implementations from \citep{cuturi2022optimal}.  The momentum hyper-parameter $\omega$ was set at $1.05$ from a grid search of $\{0.8, 1.05, 1.1, 1.3\}$. Adaptive momentum consists of adjusting the momentum parameters every \textit{adapt\_iters} number of iterations where \textit{adapt\_iters} was set to $10$ from a search on $\{10, 20, 50, 200\}$. $\epsilon$ decay consisted of gradually reducing the regularization term from $5 \epsilon$ to $\epsilon$ by a factor of $0.8$, from a search of decay factors from $\{0.8, 0.95\}$. The Anderson acceleration parameter was set to $5$ from a search on $\{3, 5, 8, 10, 15\}$.

\subsection{Soft Error Details}\label{sec:app_soft_err}
Regularization $\epsilon=0.01$ was used for the soft-error task. The soft 0/1 error objective described in \citep{cuturi2019differentiable} was used, with a neural network classifier consisting of two CNN blocks with $32$ and $64$ features respectively, and a hidden layer of hidden size 512. Each CNN block consists of two CNN layers with $3\times 3$ kernel, relu activations between CNN layers and a max pooling layer at the end of each block. Implementation including neural network architecture was taken from \citep{cuturi2022optimal}\footnote{\url{https://github.com/ott-jax/ott/tree/main/ott/examples/soft_error}}. Our proposed method was compared to other acceleration baselines using the same grid of hyperparameters as described in \S\ref{sec:app_softsort}. Batch size was set to $64$ and learning rate $0.001$.

\subsection{Differentiable  Clustering Details}\label{sec:app_clus}

The experiment was repeated for $\epsilon=0.1$ and $\epsilon=0.01$ and again compared to other acceleration baselines using the same grid of hyperparameters as described in \S\ref{sec:app_softsort}. Batch size was set to $256$ and learning rate $0.001$.

Latent dimension was set to $d_z=32$ and MNIST \citep{deng2012mnist} images are of size $d_x=28\times 28$.The decoder $D_\theta: \mathbb{R}^{d_x} \rightarrow \mathbb{R}^{2 \times d_z}$ consists of 4 hidden $[512, 512, 256, 256]$ followed by a final linear layer converting the outputted embedding to a vector of dimension $784$.
The encoder $E_\theta: \mathbb{R}^{d_x} \rightarrow \mathbb{R}^{2 \times d_z}$ consists of 4 hidden layers of depths $[512, 512, 256, 256]$ with relu activations, the final embeddings is mapped to $\mathbf{m}_i \in \mathbb{R}^{d_z}$ and $logvar_i\in \mathbb{R}^{d_z}$ by two separate linear layers without activations, where $\sigma_i=\exp{(0.5 \times logvar_i)}$. For batch $(x_i)_i$, the standard VAE loss $\ell^{ae}(\theta) = \sum_i ||x_i-\tilde{x}_i||_2^2 - 0.5 \sum_i (1+2*\log(\sigma_i)-\mathbf{m}_i^2-\sigma_i^2)$. Recall $\tilde{x}_i=D_\theta(z_i)$ and $z_i=\mathbf{m}_i+\sigma_i u_i$, $u_i\sim \mathcal{N}(\mathbf{0}_{d_z}, \mathbf{I}_{d_z})$.

    \begin{figure}
        \centering
        \includegraphics[trim=0 55 0 0,clip,width=0.3\linewidth]{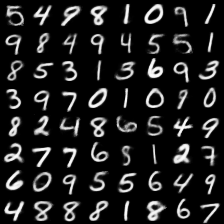}
        \caption{Generated Samples}
        \label{fig:mnist}
    \end{figure} 

    As discussed in \citep{genevay2019differentiable}, clusters may be used as an unsupervised classifier and accuracy is reported in \Cref{tab:diff_classification}, illustrating that the clusters are meaningful. In addition, samples from the clustered latent space may be used to generate new samples as a form of conditional generation, again shown in \Cref{fig:mnist}.
    
Accuracy for each cluster is defined as in \citep{genevay2019differentiable}, as follows. Accuracy for label $l$ in cluster $k$ is by $acc_{l,k}= \frac{\sum_i\mathbf{I}_{y_i==l, \tilde{y}_i==k}}{\sum_i\mathbf{I}_{y_i==k}}$ where $\tilde{y}_i = \arg\min_k ||z_i-\mu_{\phi,k}||^2_2$ and $y_i$ is the true label of $x_i$. We write the top label accuracy for each cluster $k$ as $\max_l acc_{l,k}$. When using $10$ clusters for $10$ labels for MNIST, each cluster's top label accuracy corresponds to a different label, one cluster for each digit. \Cref{tab:diff_classification} shows that the clusters manage to capture geometrically meaningful information corresponding to each label.
{
\begin{table}[ht]
    \caption{Evaluation Accuracy of trained clustered VAE for MNIST}
    \label{tab:diff_classification}
    \centering
    \begin{tabular}{lllllllllll}
    \hline
        Digit & 0 & 1 & 2 & 3 & 4 & 5 & 6 & 7 & 8 & 9 \\ 
        Accuracy & 0.91 & 0.66 & 0.42 & 0.56 & 0.80 & 0.61 & 0.68 & 0.64 & 0.90 & 0.78 \\ \hline
    \end{tabular}
\end{table}
}

\section{Overhead Analysis}\label{sec:overhead}
Although timings are highly dependent on hardware and implementation, we provide some experimental examples running on a single V100 GPU and 4 CPUs. This shows that the time overhead for DualSort and Gaussian initializers are inconsequential relative to speed-up in terms of both time and iteration count for the savings in Sinkhorn iterations. The Gaussian mixture model (GMM) is computationally more expensive than the other proposed initializers, however the table below shows that it can also result in time savings.

\subsection{Differentiable Sorting}
\begin{table}[H]
\caption{Average time in seconds for DualSort with 3 iterations and Sinkhorn iterations to convergence over 200 soft sorting problems of dimension $n$}
\centering
\begin{tabular}{llcc}
\centering
\textbf{$n$} & \textbf{Initializer} & \textbf{Initialization}                              & \textbf{Iterations} \\
 \midrule
$32$         & $\mathbf{0}$         &     -                                                 & $0.28$                 \\
             & DualSort             & $0.0012$ & $0.22$                 \\
              \midrule
$64$         & $\mathbf{0}$         &     -                                                 & $0.22$                 \\
             & DualSort             & $0.0012$                                             & $0.088$                 \\
              \midrule
$128$        & $\mathbf{0}$         &         -                                             & $0.17$                 \\
             & DualSort             & $0.0012$                                             & $0.066$                 \\
              \midrule
$256$        & $\mathbf{0}$         &     -                                                 & $0.17$                 \\
             & DualSort             & $0.0012$                                             & $0.049$                 \\
              \midrule
$512$        & $\mathbf{0}$         &      -                                                & $0.13$                 \\
             & DualSort             & $0.0012$                                              & $0.050$                \\
              \midrule
$1024$       & $\mathbf{0}$         &       -                                               & $0.14$                 \\
             & DualSort             & $0.0012$                                              & $0.058$            \\
             \bottomrule
\end{tabular}
\end{table}
It can be seen that the DualSort initialization procedure is extremely efficient and does not have significant impact on the total run-time. The timings above are averaged per OT problem over 200 runs with different seeds.

\subsection{Gaussian and GMM}
In this section we consider timings for the word embedding/ document similarity experiment. 

For the GMM initializer, the \textit{pre-compute} is the average time to compute each GMM (1 per document), divided by the number of OT problems. Each GMM is reused multiple times, so the cost is split. Each GMM was computed using scikit-learn \citep{scikit_learn} on CPU, for lack of a convenient GPU implementation. There exists open-source GPU implementations \footnote{\url{https://github.com/borchero/pycave}} of Gaussian mixture models for diagonal component covariance matrices which are significantly faster, and may be worth further investigation for more efficient implementation. Similarly, one may amortize inference in GMMs or provide a warm-start from a pooled GMM to initialize fitting the GMM. We use the default K-means initializer from scikit learn. The \textit{Initialization} field reports the time to compute the approximate dual potentials given the GMM parameters.

For the Gaussian initializer, the mean and variance parameters are inexpensive to compute, hence were not computed and cached but instead computed repeatedly on the fly for each OT problem. Hence the total initialization compute time is reported in the \textit{Initialization} column. Further computational savings could be made by caching the Gaussian parameters for each document. Note that the dimension for the Gaussian OT approximation is $d=50$ and given the Gaussian initialization is negligible here, it would also be negligible for lower dimensional settings.
\begin{table}[H]
\caption{Time, in seconds, per OT problem split by task, averaged over $1,225$ OT problems, from each pair of $50$ documents from the Newsgroup 20 dataset with a subset of vocabulary of size $n$.}
\centering
{\small{
\begin{tabular}{llllll}
\textbf{$n$}              & \textbf{Initializer} & \textbf{Pre-compute}  & \textbf{Initialization} & \textbf{Sinkhorn Iter.} & \textbf{Total} \\

 \midrule
\multirow{5}{*}{$2,000$}  & $\mathbf{0}$         & -                     & -                             & $0.059$         & $0.059$     \\
                          & Subsample            & -                     & $0.016$                      & $0.051$          & $0.067$   \\
                          & Gaus                 & -                     & $0.0028$                      & $0.045$         & $0.048$   \\
                          & GMM $K=10$           & $0.0027$              & $0.023$                       & $0.047$         & $0.073$        \\
                          & GMM $K=25$           & $0.0037$              & $0.026$                       & $0.035$         & $0.065$          \\
                          & GMM $K=50$           & $0.0047$              & $0.033$                       & $0.027$         & $0.063$          \\
                           \midrule
\multirow{5}{*}{$5,000$}  & $\mathbf{0}$         & -                     & -                             & $0.28$         & $0.28$       \\
                          & Subsample            & -                     & $0.048$                      & $0.15$         & $0.20$   \\
                          & Gaus                 & -                     & $0.0036$                      & $0.22$         & $0.22$       \\
                          & GMM $K=10$           & $0.0035$      & $0.013$                       & $0.23$         & $0.24$       \\
                          & GMM $K=25$           & $0.0070$     & $0.030$                       & $0.18$         & $0.22$      \\
                          & GMM $K=50$           & $0.012$      & $0.035$                       & $0.13$         & $0.17$    \\
                           \midrule
\multirow{5}{*}{$10,000$} & $\mathbf{0}$         & -                     & -                             & $1.05$          & $1.05$     \\
                          & Subsample            & -                     & $0.082$                      & $0.45$         & $0.53$   \\
                          & Gaus                 & -                     & $0.0053$                      & $0.81$         & $0.81$      \\
                          & GMM $K=10$           & $0.0042 $     & $0.013$                       & $0.86$         & $0.88$       \\
                          & GMM $K=25$           & $0.012 $      & $0.019$                       & $0.70$         & $0.73$     \\
                          & GMM $K=50$           & $0.022    $   & $0.035$                       & $0.56$         & $0.62$      \\
                          \bottomrule
\end{tabular}
}}
\end{table}

\section{Gaussian Potential}
In this section we derive explicitly the Gaussian potential. The transport map $T$ solving the Monge problem~\eqref{eq:monge} from a non-degenerate Gaussian measure $\mu =\mathcal{N}(\bm_\mu,\bSig_\mu)$ to another Gaussian $\nu= \mathcal{N}(\bm_\nu, \bSig_\nu)$ can be recovered in closed-form as $T^\star(x):= \mathbf{A}(x-\bm_\mu) + \bm_\nu$, where $\mathbf{A}=\bSig^{-\frac{1}{2}}_\mu(\bSig^{\frac{1}{2}}_\mu\bSig_\nu\bSig^{\frac{1}{2}}_\mu)^\frac{1}{2}\bSig^{-\frac{1}{2}}_\mu$, see e.g. \cite[Chapter 2.6]{peyre2019computational}  for a discussion. Brenier's theorem \citep{brenier1987decomposition} states that for cost $c: (x,y) \rightarrow \frac{\|x-y\|^2}{2}$ this map is uniquely defined as the gradient of a convex function $\varphi$, and it can be verified that $T^\star(x)=\nabla \varphi(x)$ where $\varphi(x)=\frac{1}{2} (x-m_\mu)^T\bA(x-m_\mu)+m_\nu^T x$.

The convex function $\varphi(x)$ is related to dual potential $f$ through $\varphi(x)=\frac{||x||^2}{2} -f(x)$ hence
\[f^\star(x) = \frac{||x||^2}{2} - \frac{1}{2} (x-m_\mu)^T\bA(x-m_\mu)
-  m_\nu^T x .\]

For cost $c: (x,y) \rightarrow \|x-y\|^2$, the optimal potential is therefore
\[f^\star(x) = ||x||^2 -  (x-m_\mu)^T\bA(x-m_\mu)
-  2m_\nu^T x .\]

\section{Convergence of Sorting Initializer and DualSort Details}\label{sec:dualsort_conv}

{
\subsection{Proof of Primal Dual Convergence}

Recovering optimal dual potentials corresponding to the primal solution is equivalent to finding any vector of shortest paths $\bff$ from a single node e.g. node 1, in the network to each of the other nodes, see e.g. \citep[Theorem 7.17]{bertsimas1997introduction} and \citep[Chapter 9]{ahuja1988network}.

 \Cref{algo:primal_init} computes the shortest path using a particular case of a method known as \textit{label correcting} \citep[Chapter 7]{bertsimas1997introduction}. Given there are no cycles, the proposed method recovers the shortest path by \citep[Theorem 7.18]{bertsimas1997introduction} and hence recovers the optimal dual potentials. 

 \Cref{algo:primal_init} exploits the primal solution efficiently by correcting all nodes in the same tree, hence the iterations are dependent on the number of trees and not necessarily the number of nodes. 
 
 The minimization step, $\bff_{s(k)} \gets \min_j c_{s(k),j}-c_{\iota(j), j}+\bff_{\iota(j)}$ follows traditional label correcting methods. However, a key insight is updating nodes along tree of $s(k)$ is equivalent to updating the minimum path to each node in the tree. 

$\bff_i$ is the shortest path to node $i$ if $\bff_{i} \leq c_{i,j}-c_{\iota(j), j}+\bff_{\iota(j)}$ $\forall j$, which is equivalent to $\bff_i+\bg_j \leq c_{i,j}$ and may be interpreted as $\bff_i$ being less than the route to any other source node $\bff_{\iota(j)}$ then to $\bff_i$ via sink node $j$, at cost $c_{i,j}-c_{\iota(j),j}$.

\subsection{DualSort Algorithm}

The \textsc{DualSort} algorithm is given sequentially below in \Cref{algo:sorting_init}. Without loss of generality, we assume that $x_i$ is rearranged in increasing order, so that the sorting permutation  $\sigma$ is the identity. Let $\textrm{diag}$ denote the operator used to extract the diagonal of a matrix, so that $\textrm{diag}(\bC)\in\mathbb{R}^n$ and one has
$[\textrm{diag}(\mathbf{C})]_i= c_{i,i}$, and write $\mathbf{1}$ for the vector of size $n$ with all entries $1$. The inner loop can be carried out in two different ways, either using a vectorized update or looping through coordinates one at a time. These two updates are distinct, and we do observe that cycling through coordinates in Gauss-Seidel fashion converges faster in terms of total number of updates. However, that perspective misses the fact that vectorized updates utilize more efficiently accelerators from a runtime perspective. Additionally, these updates are equal to, in terms of complexity to the Sinkhorn iterations, making it easier to discuss the benefits of our initializers. For these reasons, we use the \texttt{vectorized=True} flag in our experiments.

\begin{algorithm}[h]
\caption{\textsc{DualSort} Initializer}\label{algo:sorting_init_app}
        \begin{algorithmic}[1]
            \STATE{{\bfseries Input:} Cost matrix $\mathbf{C}$, primal solution, $\mathbf{P}$, \texttt{vectorized} flag}
            \STATE{\bfseries Initialize:} $\mathbf{f} = 0$
            \WHILE{not converged}
            \IF{\texttt{vectorized}}
            \STATE{$\bff \gets \min_{\text{axis=1}} \left(\mathbf{C}-\textrm{diag}(\mathbf{C})\mathbf{1}^T+\bff\mathbf{1}^T\right)$}
            \ELSE 
            \FOR{$i\in \{1, \ldots, n\}$}
            \STATE $\bff_i \gets \left(\min_j c_{i,j}-c_{j,j}+\bff_{j}\right)$
            \ENDFOR
            \ENDIF    
            \ENDWHILE
            \STATE{{\bfseries Return} {$\mathbf{f}$}}
\end{algorithmic}
\end{algorithm}

\subsection{Number of DualSort Iterations}
\Cref{fig:1d_potential} illustrates the convergence of the DualSort algorithm when compared to the true potentials found from linear programming. Visually, from the right plot of \Cref{fig:1d_potential}, the approximate dual is close to the true dual after just one iteration. However the squared error (left plot) is still large. After $3$ iterations, the error is significantly reduced and after $10$, the error is not noticeable.

\begin{figure}[h]
    \centering
    \includegraphics[width=0.45\linewidth]{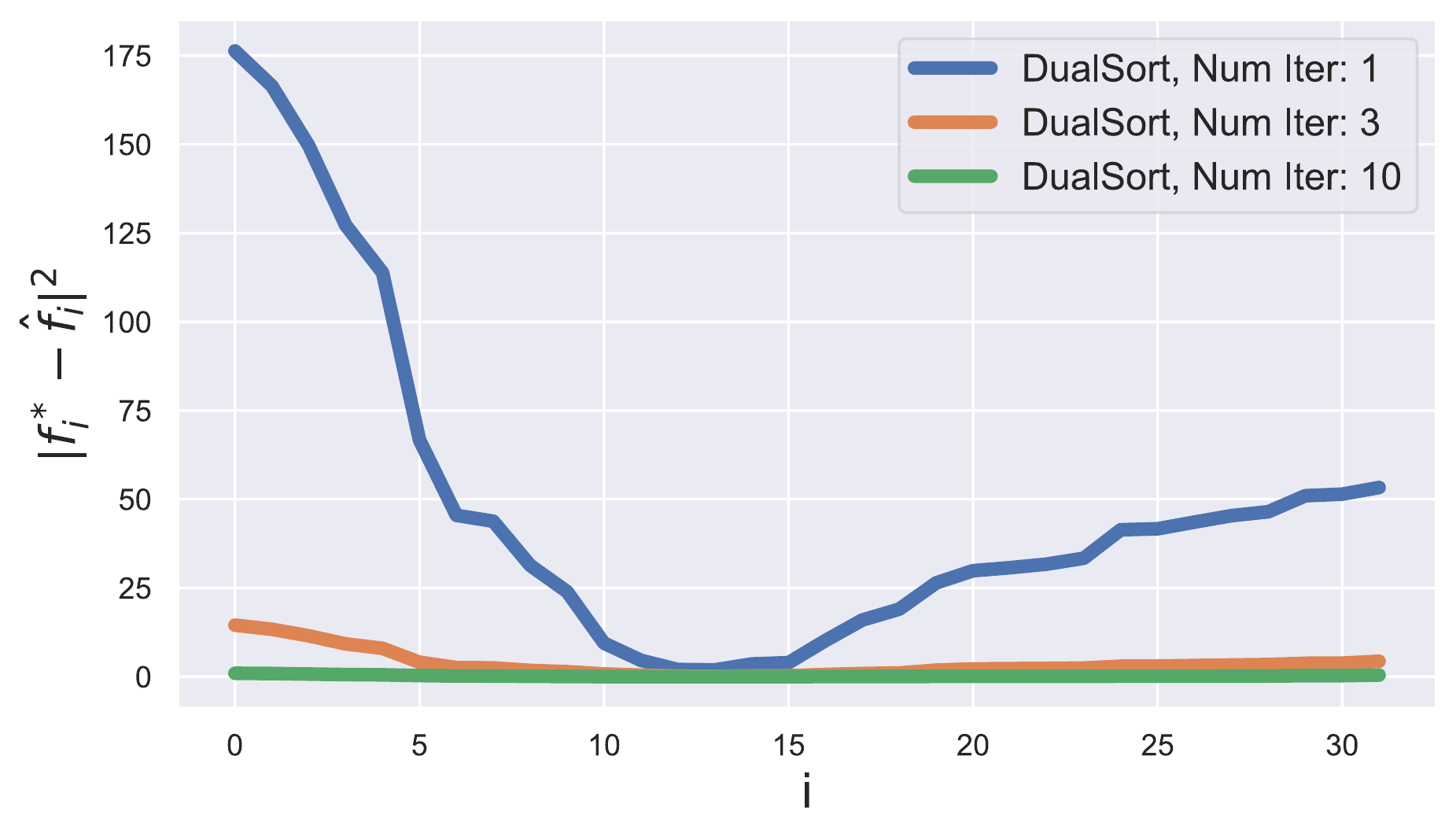}
    \includegraphics[width=0.45\linewidth]{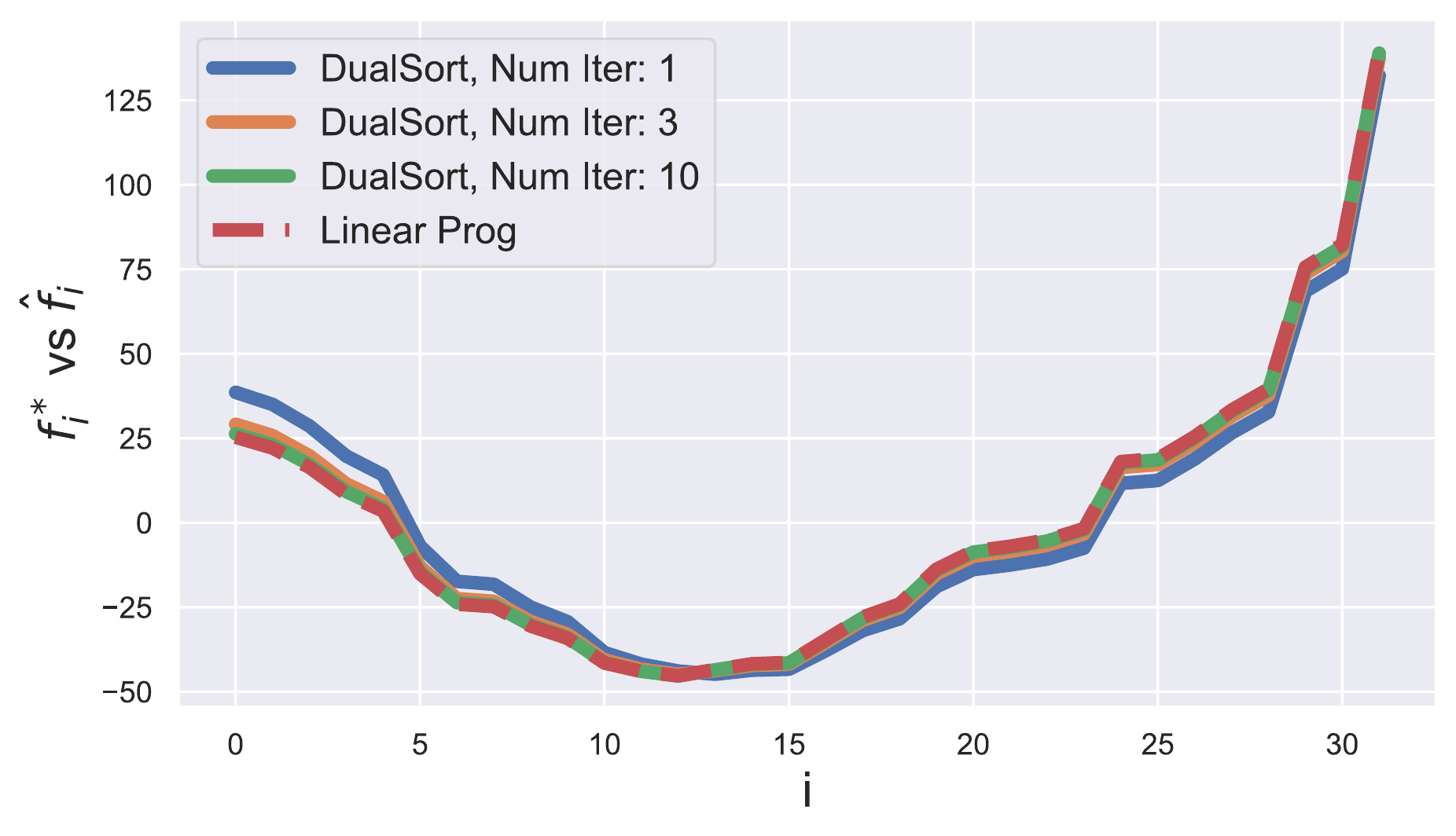}
    \caption{Single sample of size 32 from Gaussian blob dataset with 5 centers. Left: squared error vs true potential by number of DualSort iterations. Right: Potential from linear solver vs DualSort approximations.}
    \label{fig:1d_potential}
\end{figure}

\Cref{fig:thresh_maxiter} shows how the performance of the initializer improves significantly from $1$ initialization iteration to $3$ or $10$ for the CIFAR-100 soft-error classification task. Here performance is measured in how many additional Sinkhorn iterations are required after initialization for convergence. Note however that empirically there is not much difference between $3$ and $10$, hence $3$ was used in experiments.

\begin{figure}[H]
    \centering
    \includegraphics[width=\linewidth]{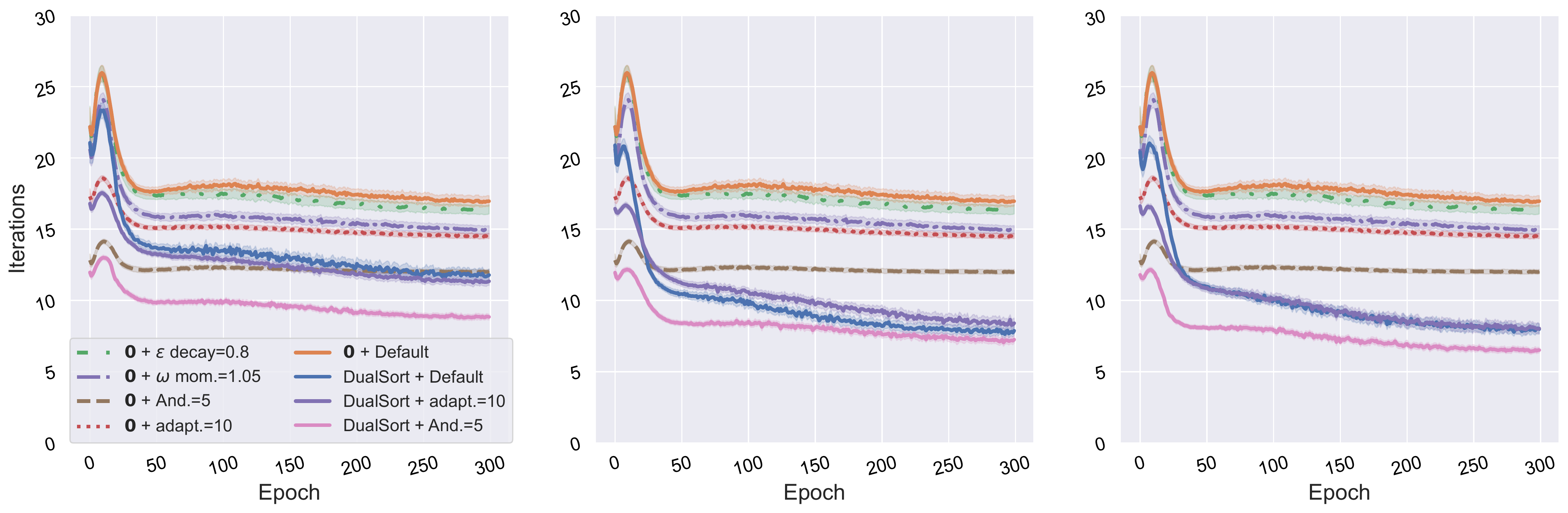}
    \includegraphics[width=\linewidth]{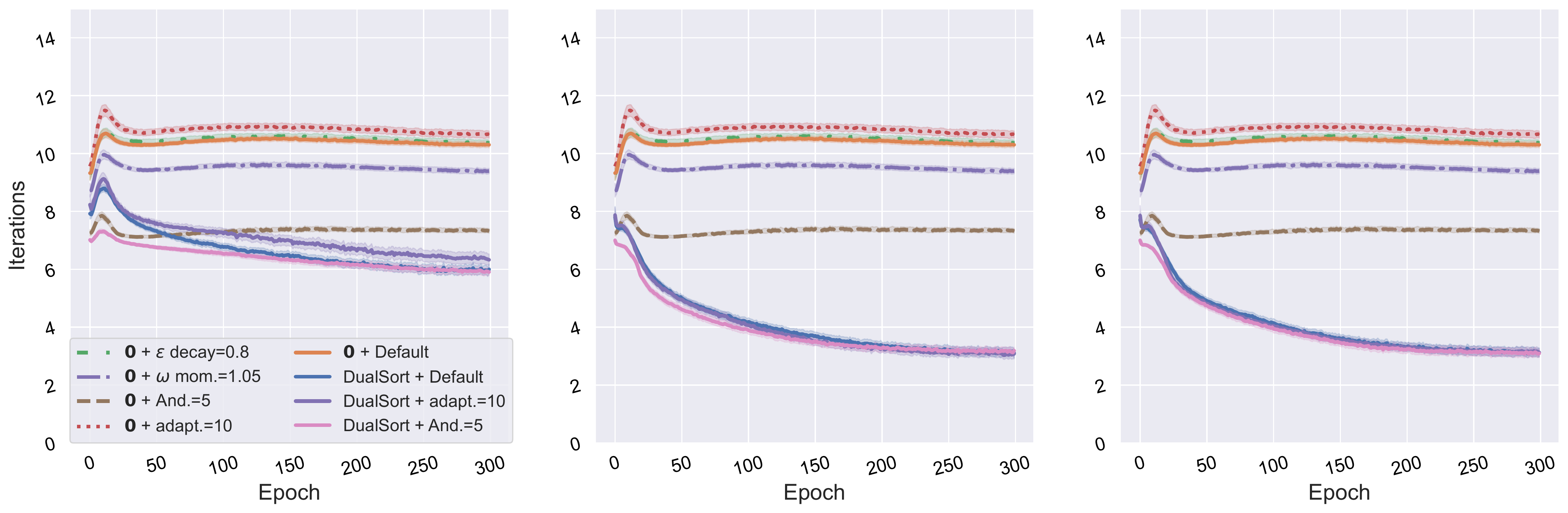}
    \caption{Number of Sinkhorn iterations per training step when using soft error loss for CIFAR-100 classifier. Top: threshold=0.01, bottom: threshold=0.05. Number of vectorized DualSort iterations 1,3,10 (left to right)}
    \label{fig:thresh_maxiter}
\end{figure}

\section{Threshold Analysis}\label{sec:app_thresh}
Convergence of each the Sinkhorn for each problem was determined according to a threshold tolerance, $\tau$, for how close the marginals from the coupling derived from potentials are to the true marginals. For OT problem between $\mu=\sum_{i=1}^na_i \delta_{x_i}$ and $\nu=\sum_{j=1}^nb_i \delta_{y_j}$, and denote potentials after $l$ Sinkhorn iterations as $\bff^{(l)}$, $\bg^{(l)}$, then the corresponding coupling may be written elementwise as $\mathbf{p}^{(l)}_{i,j}=\exp{\frac{\bff^{(l)}_i+\bg^{(l)}_j-c_{i,j}}{\epsilon}}$ and the threshold condition may be written 
\[\sum_i\lvert\sum_j \mathbf{p}^{(l)}_{i,j} -a_i\rvert + \sum_j \lvert \sum_i \mathbf{p}^{(l)}_{i,j} -b_j\rvert < \tau.\]

We use $\tau=0.01$ for speed. But also note that a higher threshold $\tau=0.05$ leads to faster convergence without drop in performance, as evidenced in \Cref{fig:app_thresh} for the soft error classification task on CIFAR-100. \Cref{fig:thresh_maxiter} also illustrates that the DualSort initializer appears to exhibit relatively better performance to the zero initialization for a higher convergence threshold. 
\begin{figure}[H]
    \centering
    \includegraphics[width=\linewidth]{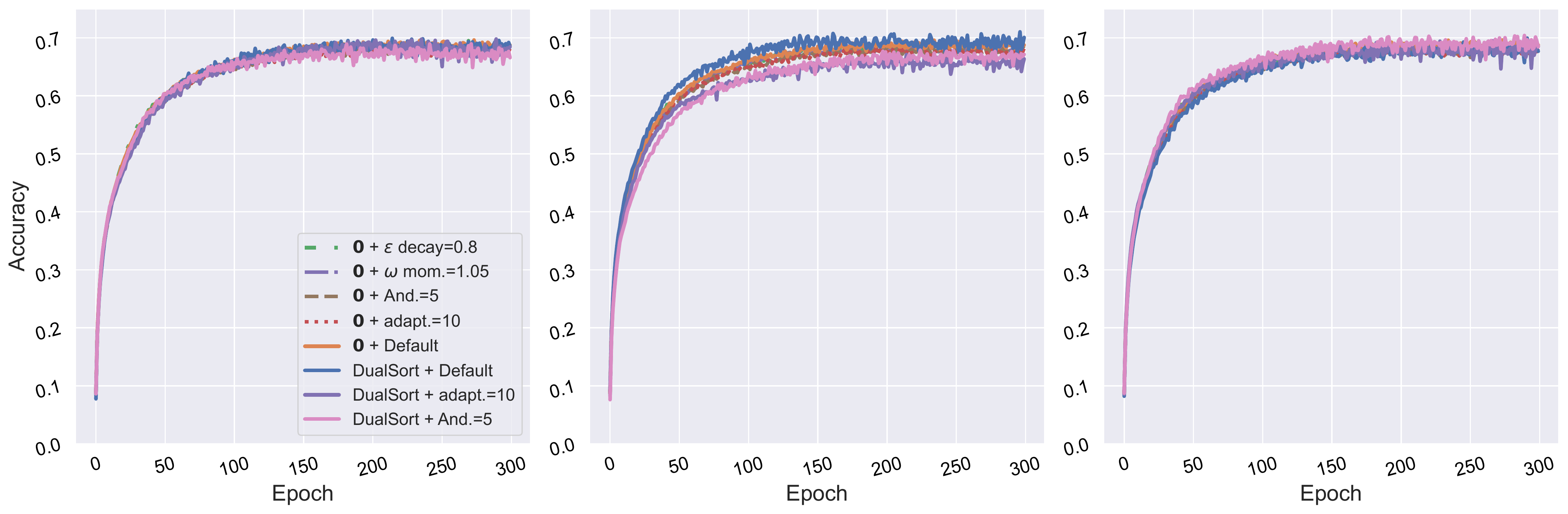}
    \includegraphics[width=\linewidth]{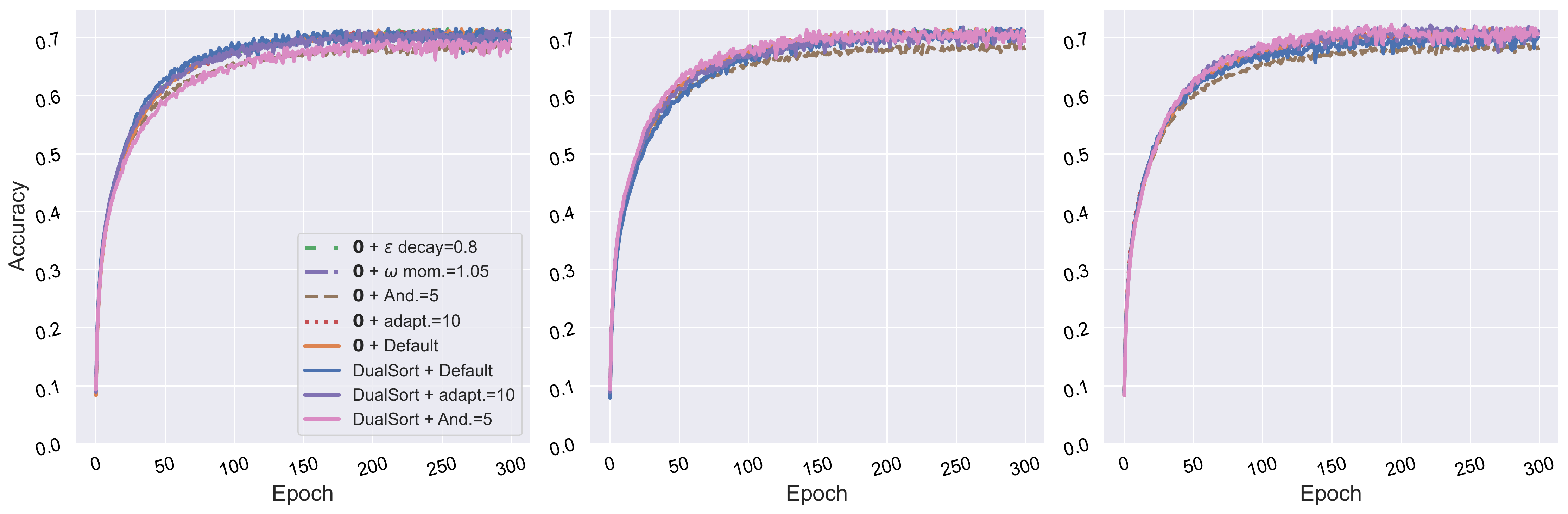}
    \caption{Evaluation accuracy through training when using soft error loss for CIFAR-100 classifier. Top: threshold=0.01, bottom: threshold=0.05. Number of vectorized DualSort iterations 1,3,10 (left to right)}
    \label{fig:app_thresh}
\end{figure}

\section{Other Details}
\textbf{Societal Impact.}  We are not aware of any direct negative societal impacts in this work. We acknowledge that the Sinkhorn algorithm may be used in various applications across compute vision and tracking with negative impacts, and this work may enable further such applications. \\
\textbf{Code.} Code for initializers has been incorporated into OTT library
\citep{cuturi2022optimal}.\\ 
\textbf{Open source software and licences.} \citep{cuturi2022optimal} has an Apache licence. 

}

\end{document}